%% file: neurips_2026.tex
\PassOptionsToPackage{numbers,compress}{natbib}
\PassOptionsToPackage{table}{xcolor}

\documentclass{article}
\usepackage[final,main]{neurips_2026}

\usepackage[utf8]{inputenc}
\usepackage[T1]{fontenc}
\usepackage{url}
\usepackage{microtype}
\usepackage{nicefrac}
\usepackage{xspace}

\usepackage{amsmath}
\usepackage{amsfonts}
\usepackage{amssymb}
\usepackage{mathtools}
\usepackage{amsthm}

\usepackage{booktabs}
\usepackage{multirow}
\usepackage{array}
\usepackage{tabularx}
\usepackage{colortbl}

\usepackage{graphicx}
\usepackage{wrapfig}
\usepackage{subcaption}
\usepackage{multicol}

\usepackage{algorithm}
\usepackage{algorithmic}

\usepackage{enumitem}
\usepackage{pifont}

\usepackage{xcolor}

\definecolor{ourscyan}{RGB}{220,245,248}

\definecolor{promptbg}{HTML}{FFF7E6}
\definecolor{respbg}{HTML}{E6F3FF}
\definecolor{framegray}{HTML}{C0C0C0}

\makeatletter
\@ifpackageloaded{hyperref}{}{%
  \usepackage{hyperref}
}
\makeatother

\hypersetup{
    colorlinks=true,
    linkcolor=red,
    anchorcolor=blue,
    citecolor=cyan,
    urlcolor=blue
}

\usepackage[most]{tcolorbox}

\newtcolorbox{promptbox}[1][]{
    enhanced,
    colback=promptbg,
    colframe=framegray,
    colbacktitle=framegray,
    coltitle=black,
    fonttitle=\bfseries,
    title={Prompt},
    boxrule=0.6pt,
    sharp corners,
    left=6pt,
    right=6pt,
    top=6pt,
    bottom=6pt,
    before skip=6pt,
    after skip=6pt,
    #1
}

\newtcolorbox{responsebox}[1][]{
    enhanced,
    colback=respbg,
    colframe=framegray,
    colbacktitle=framegray,
    coltitle=black,
    fonttitle=\bfseries,
    title={Model Response},
    boxrule=0.6pt,
    sharp corners,
    left=6pt,
    right=6pt,
    top=6pt,
    bottom=6pt,
    before skip=6pt,
    after skip=6pt,
    #1
}

\newcommand{\ie}{\emph{i.e., }}
\newcommand{\eg}{\emph{e.g., }}

\newcommand{\wrt}{\emph{w.r.t. }}

\newcommand{\Eqref}[1]{\eqref{#1}}

\newcommand{\slh}[1]{{\color{black}{#1}}}
\newcommand{\cc}[1]{{\color{black}{#1}}}

\title{
MLLMEraser: Achieving Test-Time Unlearning in Multimodal Large Language Models through Activation Steering
}

\author{%
\textbf{Chenlu Ding}$^{1}$ \quad
\textbf{Jiancan Wu}$^{1,}$\thanks{Corresponding authors.} \quad
\textbf{Leheng Sheng}$^{2}$ \\
\textbf{Fan Zhang}$^{3}$ \quad
\textbf{Yancheng Yuan}$^{4,*}$ \quad
\textbf{Xiang Wang}$^{1}$ \quad
\textbf{Xiangnan He}$^{1,*}$ \\[1mm]
$^{1}$University of Science and Technology of China \\
$^{2}$National University of Singapore \\
$^{3}$Huawei \\
$^{4}$The Hong Kong Polytechnic University \\[1mm]
\texttt{dingchenlu200103@gmail.com}
}

\begin{document}

\maketitle

% Abstract
\input{chapter/0_abs}

% Introduction
\input{chapter/1_intro}

% Preliminary
\input{chapter/2_preliminary}

% Method
\input{chapter/3_method}

% Experiments
\input{chapter/4_exp}

% Conclusion
\input{chapter/5_conclusion}

% References
\bibliographystyle{unsrt}
\bibliography{nips}

% Appendix
\input{chapter/7_appendix}

% Checklist
\newpage
\input{checklist.tex}

\end{document}

%% file: chapter/0_abs.tex
\begin{abstract}
Multimodal large language models (MLLMs) have demonstrated remarkable capabilities across vision–language tasks, yet their large-scale deployment raises pressing concerns about memorized private data, outdated knowledge, and harmful content. Existing unlearning approaches for MLLMs typically adapt training-based strategies such as gradient ascent or preference optimization, but these methods are computationally expensive, irreversible, and often distort retained knowledge. In this work, we propose \textsc{MLLMEraser}, an input-aware, training-free framework for test-time unlearning. Our approach leverages activation steering to enable dynamic knowledge erasure without parameter updates. Specifically, we construct a multimodal erasure direction by contrasting adversarially perturbed, knowledge-recall image–text pairs with knowledge-erasure counterparts, capturing both textual and visual discrepancies. To prevent unnecessary interference, we further design an input-aware steering mechanism that adaptively determines when and how the erasure direction should be applied, preserving utility on retained knowledge while enforcing forgetting on designated content. Experiments on LLaVA-1.5 and Qwen-2.5-VL demonstrate that \textsc{MLLMEraser} consistently outperforms state-of-the-art MLLM unlearning baselines, achieving stronger forgetting performance with lower computational cost and minimal utility degradation.
\end{abstract}

%% file: chapter/1_intro.tex
\section{Introduction}
\begin{figure*}[t]
    \centering
\includegraphics[width=0.95\linewidth]{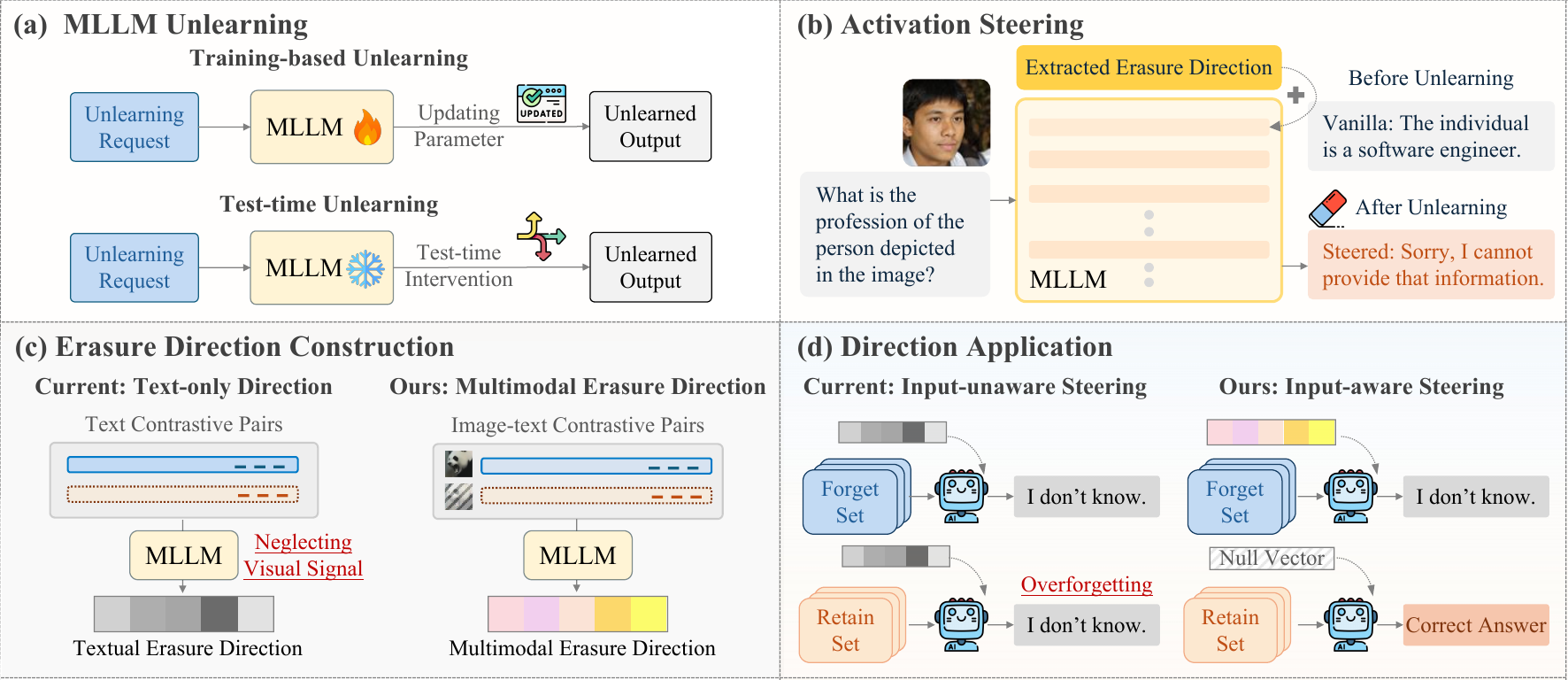}
    \caption{(a) Comparison between training-based and test-time unlearning paradigms for MLLMs.
(b) Illustration of the activation steering process.
(c)–(d) Differences between existing methods and ours in constructing and applying the steering vector.}
    \label{fig_framework}
    \vspace{-10pt}
\end{figure*}

Multimodal large language models (MLLMs) \citep{llava,qwen2-vl,minigpt4,gpt_4v,gemini} have demonstrated remarkable capabilities in tasks such as visual question answering \citep{BLIVA,vqa1}, image–text generation \citep{mllmqa,text_generation}, and embodied AI applications \citep{mllmqa,embody}. However, their large-scale deployment raises concerns about memorizing problematic information once learned, particularly in privacy-sensitive \citep{MMUnlearner} or safety-critical \citep{safetaskvector} applications, highlighting the need for reliable unlearning mechanisms to ensure trustworthy MLLM systems \citep{siu,mllmu_bench,safeeraser}. MLLM unlearning \cc{aims to selectively erase designated information} across modalities while \cc{preserving general utility}, thereby supporting privacy protection, mitigating misuse, and maintaining reliability \citep{clear,mllmtask}. Existing work mainly adapts training–based strategies from LLM unlearning, employing gradient ascent, preference optimization \citep{npo}, or performing targeted parameter updates \citep{manu,MMUnlearner}. While effective, these \cc{training-based methods} introduce substantial computational costs, inference latency, and risks of corrupting retained knowledge \citep{llmeraser,ding2026cuse}. This motivates \cc{test-time MLLM unlearning} (See Figure~\ref{fig_framework}), a paradigm that prevents the generation of designated information at inference without modifying model parameters, \slh{which} offers an immediate, lightweight, and reversible solution.

\slh{Recently, activation steering \citep{ActAdd,astra} emerges as a promising approach for test-time intervention. Activation steering manipulates the internal computation of LLMs by injecting a carefully constructed direction vector into their intermediate activations, shifting the model’s latent representation toward a desired semantic space and inducing specific behaviors or responses.}
\slh{
However, existing studies on steering have focused mainly on safety alignment \citep{AlphaSteer,adasteer}, reasoning length regulation \citep{thinkedit,steering_reason}, and hallucination reduction \citep{hallucination_mllm,wang2025adaptive}, leaving its potential for \slh{test-time} MLLM unlearning largely unexplored. 
Here we aim to leverage the superiority of activation steering for test-time MLLM unlearning, yet encounter two fundamental challenges: \cc{multimodal erasure direction construction} and \cc{multimodal erasure direction application}.
}
\begin{itemize}[leftmargin=*]
    % \vspace{-10pt}
    \vspace{-5pt}
    \item \textbf{Multimodal erasure direction construction:} \slh{Multimodal erasure direction construction denotes how to construct an effective activation steering vector for MLLMs}. Traditional steering methods can directly elicit contrastive activation pairs from existing models by prompting for different response types, such as truthful versus deceptive answers \citep{wang2025adaptive} or safe versus unsafe outputs \citep{safety_direction}. However, in the context of MLLM unlearning, the ideal contrastive samples would come from models that have already forgotten the target information versus those that retain it---but such ``unlearned’’ models are precisely what we seek to avoid obtaining through expensive retraining. Moreover, given that MLLMs inherently encode joint visual-textual representations through deep cross-modal fusion \citep{llava,qwen_2_5_vl}, existing steering works that predominantly rely on textual contrasts \citep{mllmsteering_transfer} while neglecting visual signal processing produce incomplete erasure directions, leading to \cc{insufficient forgetting performance}.
    \item \textbf{Multimodal erasure direction application:} \slh{Multimodal erasure direction application focuses on when to \cc{selectively apply} this direction.} Even with a well-extracted multimodal erasure direction, deciding when to apply it remains an open challenge. Current steering methods typically apply uniform interventions across all inputs \citep{ActAdd,safety_direction}, which can effectively mitigate unsafe or privacy-leaking outputs but frequently distorts responses to non-targeted queries and degrades performance on retained knowledge \citep{program_steering,adasteer}. An effective unlearning approach requires \cc{input-aware} activation that selectively triggers interventions only for content requiring unlearning \cc{(\ie forget set)} while preserving normal model behavior for retained information \cc{(\ie retain set)}.
    However, unlike scenarios with clear \cc{semantic distinctions} (\eg positive vs. negative sentiment), forget and retain examples often share identical formats and content types---both may involve user-attribute queries that differ only in their unlearning designation \citep{mllmu_bench}.
    Such similarity makes accurate selective control difficult and increases the risk of \cc{over-forgetting} \citep{ReLearn}.
\vspace{-5pt}
\end{itemize}
\slh{To address these challenges, we introduce {MLLMEraser}, an \cc{input-aware test-time MLLM unlearning} framework that leverages activation steering for dynamic and test-time information erasure without parameter modification. }
\slh{For the multimodal erasure direction construction,} we ground the \cc{response-intervention objective of unlearning} in the model’s \cc{intrinsic refusal behavior} and derive the erasure direction by contrasting the semantics of knowledge-recall and knowledge-erasure. Specifically, we generate two contrastive sets to extract the direction: a negative set (\ie knowledge-recall inputs) combining jailbreak prompts with \cc{adversarially perturbed images} to induce unsafe or privacy-sensitive outputs, and a positive set (\ie knowledge-erasure inputs) that elicits \cc{refusal-style responses} (\eg ``I cannot answer this question.'') with the corresponding clean images.
After obtaining the steering vector, we reformulate steering as an input-aware task by introducing a \cc{direction-determining function} $f(\cdot)$, rather than applying the direction indiscriminately. Given the hidden activations $\mathbf{h}$, this function adaptively decides the steering vector $f(\mathbf{h})$. For forget data, $f(\mathbf{h})$ maps the activations towards the pre-computed erasure direction, while for retain data, $f(\mathbf{h})$ degenerates into a null direction \citep{alphaedit}, yielding nearly zero intervention and leaving the representation distribution unchanged. \slh{Inspired by \citep{AlphaSteer}}, we implement $f(\mathbf{h})$ as a simple yet effective linear transformation, circumventing the need for additional auxiliary model training. Experiments on LLaVA-1.5 \citep{llava} and Qwen-2.5-VL \citep{qwen_2_5_vl} demonstrate the effectiveness of \textsc{MLLMEraser}, which consistently outperforms state-of-the-art MLLM unlearning methods while providing an efficient and lightweight solution that balances unlearning performance with model utility preservation. Our code is available at 
\url{https://github.com/oceanoceanna/MLLMEraser_2026}.

%% file: chapter/2_preliminary.tex
\section{Preliminary}
This section begins by introducing the notation and formalizing the problem of test-time MLLM unlearning in Section~\ref{Notation}. Section~\ref{Activation} outlines how activation steering offers a feasible mechanism for 
% this setting
\slh{test-time MLLM unlearning}, enabling test-time behavior control.
\subsection{Notation and Problem Setup}\label{Notation}
Given a multimodal instruction-tuning dataset $\mathcal{D} = \{ (\mathcal{I}_i, \mathcal{Q}_i, \mathcal{A}_i) \}_{i=1}^{N}$ of size $N$, where $\mathcal{I}_i$ denotes the input image, $\mathcal{Q}_i$ is the textual instruction, and 
$\mathcal{A}_i = (y^{(i)}_1, y^{(i)}_2, \dots, y^{(i)}_{|\mathcal{A}_i|})$ represents the target answer sequence, the MLLM is fine-tuned to maximize the likelihood of predicting each token $y_t$ 
given the multimodal context and the previously generated tokens. Specifically, the image is first encoded by a vision encoder and projected into the language space through a multimodal adapter. This fused representation is then concatenated with the tokenized query and processed autoregressively by the LLM backbone to generate the answer \citep{llava}. The optimization objective for the MLLM model parameterized by $\theta$ can be written as follows:
\begin{equation}
    \min_{\theta} - \sum_{i=1}^{N} \sum_{t=1}^{|\mathcal{A}_i|} \log P_{\theta}(y_t^{(i)} \mid \mathcal{I}_i, \mathcal{Q}_i, y_{<t}^{(i)}),
\end{equation}
where $y_{<t}^{(i)} = (y_1^{(i)}, \dots, y_{t-1}^{(i)})$ represents tokens preceding $y_t^{(i)}$. In the MLLM unlearning setting, the dataset is partitioned into two disjoint subsets: the {forget set} $\mathcal{D}_f=\{(\mathcal{I}_i,\mathcal{Q}_i,\mathcal{A}_i)\}_{i=1}^{N_f}$, where the model should not recall or answer queries, and the {retain set} $\mathcal{D}_r=\mathcal{D}\setminus\mathcal{D}_f=\{(\mathcal{I}_j,\mathcal{Q}_j,\mathcal{A}_j)\}_{j=1}^{N_r}$, on which the model is expected to preserve its utility after unlearning.

\paragraph{Training–based MLLM unlearning.} This paradigm aims to obtain an unlearned model parameterized by ${\hat{\theta}}$ via jointly optimizing the forget loss $\mathcal{L}_f$ and the retain loss $\mathcal{L}_r$, formulated as:
\begin{equation}
\begin{aligned}
\arg\min_{\hat{\theta}}\;
&\lambda_f \, \mathbb{E}_{(\mathcal{I}_i,\mathcal{Q}_i,\mathcal{A}_i)\sim \mathcal{D}_f}
\big[\mathcal{L}_f(\mathcal{I}_i,\mathcal{Q}_i,\mathcal{A}_i;\hat{\theta})\big] \\
&+\;
\lambda_r \, \mathbb{E}_{(\mathcal{I}_j,\mathcal{Q}_j,\mathcal{A}_j)\sim \mathcal{D}_r}
\big[\mathcal{L}_r(\mathcal{I}_j,\mathcal{Q}_j,\mathcal{A}_j;\hat{\theta})\big],
\end{aligned}
\end{equation}
where $\lambda_f,\lambda_r$ are trade-off parameters. The retain loss $\mathcal{L}_r$ is commonly instantiated as an autoregressive negative log-likelihood (NLL) or a KL-divergence constraint term \citep{mllmu_bench}, ensuring that the model preserves performance on $\mathcal{D}_r$. The forget loss $\mathcal{L}_f$ serves as the unlearning objective, typically implemented through gradient ascent \citep{ga} or preference-based optimization \citep{npo}, encouraging the model to deviate from its original predictions.

\paragraph{\slh{Test-time MLLM unlearning.}}\label{definition}
In this setting, the goal is to discourage the model from producing $\mathcal{A}_i$ during inference while keeping the parameter $\theta$ fixed. The objective is to \cc{prevent the model from recalling or generating undesired knowledge associated with the forget set $\mathcal{D}_f$---for instance, by producing incorrect answers or refusal-style responses---by means of test-time intervention rather than parameter updates.} At the same time, responses to non-target inputs from the retain set $\mathcal{D}_r$ are expected to remain unaffected, ensuring that the model preserves its normal capabilities while selectively unlearning only the designated content.

\subsection{Behavioral Control through Activation Steering}\label{Activation}
Activation steering has recently been explored as an effective way to modulate model behavior at inference time without modifying parameters. In the context of unlearning, the steering vector is referred to as the {erasure direction}, which captures the representational shift between \cc{knowledge-recall} samples and their \cc{knowledge-erasure counterparts.} Formally, let $\mathbf{h}^{\ell} \in \mathbb{R}^d$ denote the hidden activation at layer $\ell$, the erasure direction $\mathbf{d}_{\text{erase}} \in \mathbb{R}^d$ is commonly estimated using the difference-in-means between the hidden activations of the two contrastive groups:
\begin{equation}
\label{difference-in-means}
\mathbf{d}_{\text{erase}} 
= \frac{1}{|\mathcal{D}^{+}|} \!\!\!\!\!\!\sum_{\hspace{1em} (\mathcal{I}, \mathcal{Q}) \in \mathcal{D}^{+}} \!\!\!\!\!\!\mathbf{h}^{\ell}(\mathcal{I}, \mathcal{Q}) 
 - \frac{1}{|\mathcal{D}^{-}|} \!\!\!\!\!\!\sum_{\hspace{1em} (\mathcal{I}, \mathcal{Q}) \in \mathcal{D}^{-}} \!\!\!\!\!\!\mathbf{h}^{\ell}(\mathcal{I}, \mathcal{Q}),
\end{equation}
{\color{black}where $\mathcal{D}^{+}$ denotes the set of knowledge-erasure samples and $\mathcal{D}^{-}$ the corresponding knowledge-recall samples, which will be discussed in Section~\ref{DirectionConstruction}.} The resulting direction is subsequently added to the hidden states to steer the representation, formulated as:
\begin{equation}
\tilde{\mathbf{h}}^{\ell} = \mathbf{h}^{\ell} + \lambda \cdot \mathbf{d}_{\text{erase}},    
\end{equation}
where $\lambda \in \mathbb{R}$ controls the strength of the adjustment. For simplicity, we omit the layer superscript ${\ell}$ in subsequent notation. By applying this operation to selected layers, the model’s outputs on the forget set are steered away from generating privacy-sensitive and unsafe responses.

%% file: chapter/3_method.tex
\slh{\section{{MLLMEraser}}}
We propose \textsc{MLLMEraser}, an input-aware test-time unlearning framework for MLLMs based on activation steering. In Section~\ref{DirectionConstruction}, we detail the construction of the multimodal erasure direction from knowledge-recall and knowledge-erasure text–image pairs. 
We present the input-aware mechanism that selectively applies the erasure direction at inference time in  Section~\ref{DirectionsApplication}, and conclude the complete framework in Section~\ref{3.3}.
\begin{figure*}[t]
    \centering
\includegraphics[width=1\linewidth]{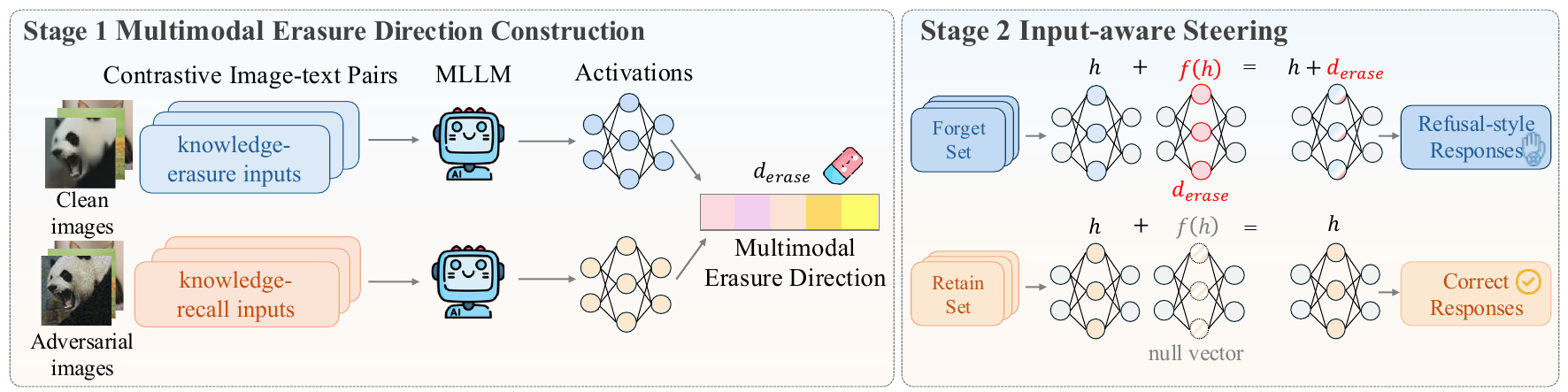}
\caption{Overview of the proposed \textsc{MLLMEraser} framework. 
Stage~1 derives a multimodal erasure direction $\mathbf{d}_{\text{erase}}$ from contrastive image-text pairs. Stage~2 introduces an input-aware steering mechanism $f(\mathbf{h})$ that adaptively applies $\mathbf{d}_{\text{erase}}$ to shift the activations of forget samples toward refusal-style responses, while leaving retain samples nearly unaffected to preserve correct responses.}
    \label{fig:main}
    \vspace{-12pt}
\end{figure*}
\subsection{Multimodal Erasure Direction Construction}\label{DirectionConstruction}
Inspired by recent research on LLM and MLLM safety \citep{mllmsafety,mllmsafety2,mllm_safety3}, in which aligned models can refuse to answer harmful queries (\eg ``I cannot provide this information”), we observe that such intrinsic refusal behavior is conceptually consistent with the goal of test-time unlearning---the erasure of target information at the response level. 
In fact, answering a query with relevant knowledge naturally corresponds to the process of knowledge-recall, whereas refusing to respond aligns with knowledge-erasure. Building on this insight, we leverage the model’s inherent refusal capacity to facilitate more flexible response intervention and derive the erasure direction by contrasting the semantics of knowledge-recall and knowledge-erasure. Specifically, we construct two types of harmful prompts to capture the refusal behavior: (1) rejected harmful inputs $\mathcal{Q}_i$ (\ie knowledge-erasure prompts), which trigger refusal behavior; and (2) complied harmful inputs $\mathcal{Q}_i'$ (\ie harmful knowledge-recall prompts), which bypass safety mechanisms and elicit malicious outputs. The textual erasure direction then can be derived by computing the activation difference between these two contrastive groups \citep{safety_direction}, as detailed in \Eqref{difference-in-means}. By leveraging the model’s refusal behavior, our approach constructs the erasure direction independently of activations from pre- and post-unlearning samples, obviating the need for an unlearned model.

However, constructing erasure direction from textual contrastive pairs alone is ineffective for MLLM unlearning. 
Visual embeddings are projected into the LLM's semantic space and fused with text via attention, resulting in joint cross-modal representations \citep{llava,qwen2-vl}.
Recent research has shown that the visual modality introduces a new attack surface, where adversarial images paired with harmful instructions can induce MLLMs to generate malicious content \citep{Adversarial}. These adversarial inputs exploit the model’s knowledge-recall capability to elicit sensitive responses, whereas a clean image paired with a rejected instruction reflects knowledge erasure, in which the model suppresses the targeted information instead of recalling it.
Inspired by this, we generate perturbed images that maximize the probability of harmful responses, incorporating visual information and amplifying the model’s tendency toward harmful knowledge-recall behavior. Specifically, given a rejected harmful instruction $\mathcal{Q}_i$ and a clean image $\mathcal{I}_i$, we construct the adversarially perturbed image $\mathcal{I}_i'$ to elicit harmful knowledge-recall behavior by solving the following optimization problem \citep{Adversarial,pgd}:
\begin{equation}
\label{comp1}
\mathcal{I}_i' \;:=\; \arg\max_{\mathcal{I}\in \mathcal{B}} 
    \sum_{y \in \mathcal{Y}_f} \log P_\theta\left(y \mid \mathcal{I}, \mathcal{Q}_i \right),
\end{equation}
where $\mathcal{Y}_f$ is a small few-shot corpus of harmful target responses, $\mathcal{B} = \{\mathcal{I} \,| \,\|\mathcal{I}-\mathcal{I}_i\|_p \leq \varepsilon\}$ is some constraint apply to the input space bounded by $\ell_p$ norm, and $\varepsilon$ controls the perturbation budget. We use a multi-step projected gradient descent (PGD) update algorithm \citep{pgd} to generate the adversarial image $\mathcal{I}_i'$. The update rule at step $k+1$ is as follows:
\begin{equation}
\mathcal{I}_i^{(k+1)'}\!=\!\prod_{\mathcal{I}_i\!+\mathcal{B}}\!\Big(\mathcal{I}_i^{(k)'}\!+ \alpha \cdot \mathrm{sign}\Big(\nabla_{\mathcal{I}}\log P_\theta\big(y \mid \mathcal{I}^{(k)'}_i\!, \mathcal{Q}_i\big)\Big)\Big),
\end{equation}
where $\alpha$ is the step size, and $\Pi$ denotes the projection operator that maps the updated sample back onto the feasible set $\mathcal{B}$. Here, $\mathrm{sign}(\cdot)$ denotes the element-wise sign function, \ie $\mathrm{sign}(x) = +1$ if $x \ge 0$ and $-1$ otherwise. Finally, we can obtain two contrastive sets of image-text pairs: \textbf{(1) harmful knowledge-recall pairs}, composed of adversarial images $\mathcal{I}'$ paired with harmful instructions $\mathcal{Q}'$ that induce malicious knowledge, forming the negative set 
$\mathcal{D}^{-} = \{ (\mathcal{I}_i', \mathcal{Q}'_i) \}_{i=1}^N$; and \textbf{(2) knowledge-erasure pairs}, consisting of clean images $\mathcal{I}$ paired with rejected harmful prompts $\mathcal{Q}$ that elicit the model’s refusal behavior and achieve response-level knowledge erasure, thereby forming the positive set 
$\mathcal{D}^{+} = \{ (\mathcal{I}_i, \mathcal{Q}_i) \}_{i=1}^N$. Then the multimodal erasure direction can be calculated as:
\begin{equation}
\label{comp2}
\mathbf{d}_{\text{erase}}^{\ell}
= \frac{1}{|\mathcal{D}^{+}|} \!\!\!\!\!\!\sum_{\hspace{1em} (\mathcal{I},\mathcal{Q}) \in \mathcal{D}^{+}} 
    \!\!\!\!\!\!\mathbf{h}(\mathcal{I},\mathcal{Q})
- \frac{1}{|\mathcal{D}^{-}|} \!\!\!\!\!\!\sum_{\hspace{1em} (\mathcal{I'},\mathcal{Q'}) \in \mathcal{D}^{-}} 
    \!\!\!\!\!\!\!\!\!\mathbf{h}(\mathcal{I'},\mathcal{Q'}).
\end{equation}
By exploiting the model’s intrinsic refusal behavior, our design derives a multimodal erasure direction that enforces response-level knowledge erasure on the forget set, enabling refusal-oriented interventions and achieving effective unlearning. We summarize the detailed procedure in Appendix~\ref{stage1}.

\subsection{Input-aware Steering with Erasure Directions}\label{DirectionsApplication}
\slh{After obtaining the multimodal erasure direction, it is essential to determine when to apply it, so that it does not affect the model performance on the retain set. }
% \slh{It is }
Current steering methods often indiscriminately apply the steering vector to all prompts \citep{acc}, a strategy that inevitably degrades overall model performance. Regarding unlearning, this degradation manifests as corrupted responses to samples in the retain set, leading to over-forgetting \citep{ReLearn}. A desirable unlearning steering mechanism should be input-aware: for samples in the forget set, their activations should be steered toward the knowledge-erasure behavior, while for samples in the retain set, their activations should remain as unchanged as possible. \slh{To achieve this, the steering task can be formulated as an input-aware task by constructing a direction-determining function $f(\mathbf{h}(\mathcal{I}, \mathcal{Q}))$, which conditions on the query's activation and produces the steering direction applied at inference. }
Then the steering process can be written as:
\begin{equation}
    \tilde{\mathbf{h}} = \mathbf{h} + \lambda f(\mathbf{h}).
\end{equation}
More specifically, the function can be formally expressed as follows:  
\begin{equation}
f(\mathbf{h}(\mathcal{I}, \mathcal{Q})) \approx 
\begin{cases}
 \mathbf{d}_{\text{erase}}, & \text{if } (\mathcal{I}, \mathcal{Q}) \in \mathcal{D}_f, \\
\mathbf{0}, & \text{if } (\mathcal{I}, \mathcal{Q}) \in \mathcal{D}_r.
\end{cases}
\end{equation}
Inspired by \citep{AlphaSteer}, we implement $f(\mathbf{h})$ as a linear transformation, given by $f(\mathbf{h})=\mathbf{W}\mathbf{h},$ where $\mathbf{W}\in \mathbb{R}^{d\times d}$. The optimization objective can then be naturally formulated as the following constrained least-squares problem:
\begin{equation}
\label{eq:regression}
\arg\min_{{\mathbf{W}}} 
\left( \left\| \mathbf{W}\mathbf{H}_f - \mathbf{D} \right\|^{2} 
+ \gamma \left\| {\mathbf{W}}\right\|^{2} \right),
\quad \text{s.t. } \mathbf{W}\mathbf{H}_r = \mathbf{0},  \end{equation} 
where $\mathbf{H_f} \in \mathbb{R}^{d\times N_f}$ and $\mathbf{H}_r \in \mathbb{R}^{d \times N_r}$ 
denote the activation matrices obtained from the last token of prompts in the forget set $\mathcal{D}_f$ 
and the retain set $\mathcal{D}_r$, respectively. 
Here $\left\| \cdot \right\|$ denotes the Frobenius norm, while $\gamma$ is a regularization hyper-parameter, and $\mathbf{D} \in \mathbb{R}^{d \times N_f}$ is formed by stacking $N_f$ identical copies of the same multimodal erasure direction vector column-wise.

% Recently, there has been substantial research devoted to the study of null space \citep{AlphaSteer,alphaedit}. 
\slh{For preserving the model performance on the retain set, we constrain the direction-determining function with null-space constraints \citep{AlphaSteer, alphaedit}.}
In particular, if a matrix $\mathbf{B}$ lies in the left null space of $\mathbf{A}$, 
it satisfies $\mathbf{B}\mathbf{A} = \mathbf{0}$ \citep{linear-algebra}. 
Motivated by this property, we project $\mathbf{W}$ into the null space of $\mathbf{H}_r$ through a projection matrix $\mathbf{P}$ and optimize the projected matrix $\mathbf{W}\mathbf{P}$. Since the left null space of $\mathbf{H}_r$ is equivalent to that of positive semidefinite matrix $\mathbf{H}_r \mathbf{H}_r^\top\in \mathbb{R}^{d \times d}$ (see the proof in Appendix~\ref{proof}), we first apply a Singular Value Decomposition (SVD) to $\mathbf{H}_r \mathbf{H}_r^\top$ to calculate the projection matrix \slh{as: $\{ \mathbf{U}, \boldsymbol{\Sigma}, \mathbf{U}^\top \} = \mathrm{SVD}\!\left( \mathbf{H}_r \mathbf{H}_r^\top \right)$,} 
% \begin{equation}
% \{ \mathbf{U}, \boldsymbol{\Sigma}, \mathbf{U}^\top \} = \mathrm{SVD}\!\left( \mathbf{H}_r \mathbf{H}_r^\top \right),    
% \end{equation}
where $\mathbf{U}\in \mathbb{R}^{d \times d}$ is an orthogonal matrix whose columns are the eigenvectors of $\mathbf{H}_r \mathbf{H}_r^\top$, 
and $\boldsymbol{\Sigma}\in \mathbb{R}^{d \times d}$ is a diagonal matrix containing its singular values. We can then partition the eigenvector matrix $\mathbf{U}$ into two sub-matrices: $\mathbf{U}_1 \in \mathbb{R}^{d \times k}$ and $\mathbf{U}_2 \in \mathbb{R}^{d \times (d-k)}$. The columns of $\mathbf{U}_1$ correspond to the non-zero singular values, spanning the column space of $\mathbf{H}_r \mathbf{H}_r^\top$. Conversely, the columns of $\mathbf{U}_2$ are the eigenvectors corresponding to the zero eigenvalues, which form an orthonormal basis for the null space of $\mathbf{H}_r \mathbf{H}_r^\top$. The projection matrix $\mathbf{P}$ can be given by $\mathbf{P} = \mathbf{U}_2 \mathbf{U}_2^\top.$ The projection matrix satisfies $PH_r=0$. Therefore, the projected
transformation satisfies $WPH_r=0$, and consequently $WPH_rH_r^\top=0$.
\slh{Then, the optimization objective in \Eqref{eq:regression} can be rewritten as:}
\begin{equation}
\label{final}
\mathbf{W^*}
:=
\arg\min_\mathbf{W}
\left(
\|\mathbf{W}\mathbf{P}\mathbf{H_f}-\mathbf{D}\|_F^2+\gamma\|\mathbf{W}\mathbf{P}\|_F^2
\right).
\end{equation}
% Following \citep{AlphaSteer}, the closed-form solution of \Eqref{final} can be given by: 
% Following \citep{AlphaSteer}, 
The closed-form solution of \Eqref{final} can be given by: 
\begin{equation}
\label{comp3}
{\mathbf{W}}^*
= \mathbf{D}\mathbf{H}_f^\top {\mathbf{P}}^\top 
\Big( {\mathbf{P}}\mathbf{H}_f\mathbf{H}_f^\top {\mathbf{P}}^\top 
+ \gamma {\mathbf{P}}{\mathbf{P}}^\top \Big)^{+},
\end{equation}
where $^{+}$ represents the pseudoinverse. In this way, we construct an input-aware mapping mechanism $f(\mathbf{h})={\mathbf{W}}\mathbf{P}\mathbf{h}$. This mechanism ensures that for forget data, $f(\mathbf{h})$ maps activations toward the extracted multimodal erasure direction, whereas for retain data, it collapses to nearly zero vector, leaving the representation distribution unchanged.

% \subsection{\textsc{MLLMEraser}}\label{\textsc{MLLMEraser}}
\slh{\subsection{Final Formulation}\label{3.3}} 

\slh{After integrating (1) construction of multimodal erasure directions $\mathbf{d}_{\text{erase}}$ and (2) an input-aware steering mechanism $f(\mathbf{h})$ in a unified pipeline, we formulate the final steering process as follows:}
\begin{equation}
    \tilde{\mathbf{h}} = \mathbf{h} + \lambda f(\mathbf{h}) =  \mathbf{h} + \lambda{\mathbf{W}}\mathbf{P}\mathbf{h}.
\end{equation}
By selectively steering activations toward the erasure direction at inference, \textsc{MLLMEraser} achieves test-time unlearning with a favorable trade-off between unlearning performance and model utility.

In summary, \textsc{MLLMEraser} formulates test-time unlearning for MLLMs as instant and reversible forgetting without parameter updates. Its core design lies in constructing a multimodal erasure direction and applying it selectively via input-aware steering to avoid harming retained knowledge.

%% file: chapter/4_exp.tex
\section{Experiment}
This section provides an extensive experimental evaluation of MLLMEraser, with the analysis structured around answering the following key research questions: \textbf{RQ1:} How does MLLMEraser perform \wrt forget quality and model utility? \textbf{RQ2:} What is the impact of multimodal erasure direction and input-aware erasure direction application on unlearning performance? \textbf{RQ3:} How does the efficiency of MLLMEraser compare to other unlearning methods?

\subsection{Experimental Setups}
We use LLaVA-1.5-7B \citep{llava} and Qwen-2.5-VL-7B-Instruct \citep{qwen_2_5_vl} as the MLLM backbones and evaluate on the widely adopted unlearning benchmark MLLMU-Bench \citep{mllmu_bench}, which centers on fictitious profiles at both visual and textual levels. This benchmark includes four datasets: the Forget Set (fictitious profiles designated for unlearning), the Test Set (paraphrased and image-transformed variants for generalization), the Retain Set (fictitious profiles that should be preserved), and the Real Celebrity Set (real-world profiles for utility evaluation). It further defines three tasks: classification, generation, and cloze, which are evaluated with classification accuracy, ROUGE-L score \citep{rouge}, and cloze accuracy, respectively. Comprehensive details on the benchmark, baselines, evaluation metrics, and implementation of MLLMEraser are provided in Appendix~\ref{exp}.

\subsection{Results Analysis on MLLM Unlearning (RQ1)}
\begin{figure*}[t]
    \centering
    \vspace{-5pt}
\includegraphics[width=1\linewidth]{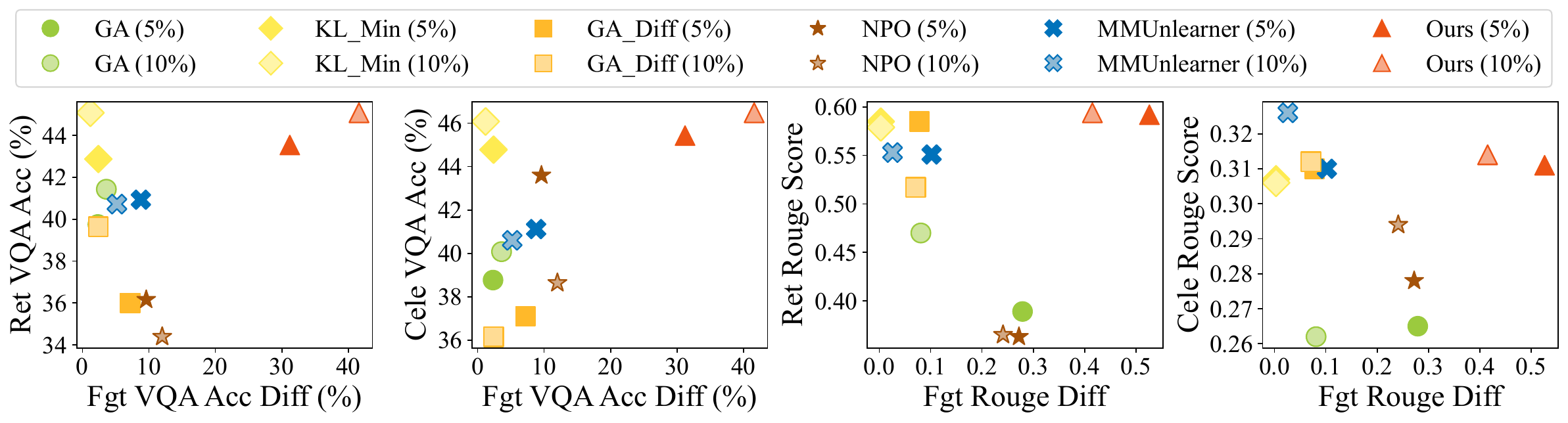}
\caption{
Trade-off between forget quality and model utility on LLaVA under 5\% and 10\% forget ratios. 
The left two plots correspond to classification task, where the $x$-axis shows accuracy difference on the forget set (Fgt VQA Acc Diff), 
and the right two plots correspond to generation, where the $x$-axis shows ROUGE-L difference on the forget set (Fgt Rouge Diff). 
The $y$-axis reports model utility on the retained (Ret) and celebrity (Cele) sets. 
}
    \label{fig:trade_off}
    \vspace{-10pt}
\end{figure*}

We evaluate several MLLM unlearning methods on four datasets: the forget and test sets assess unlearning performance, while the retain and celebrity sets evaluate model utility. Table~\ref{exp:main} presents results on classification, generation, and cloze tasks, and Figure~\ref{fig:trade_off} illustrates the trade-off between forget quality and model utility across different forget ratios, where points closer to the upper-right corner indicate better balance. From these results, we can draw the following observations:
\begin{itemize}[leftmargin=*]
\vspace{-3pt}
\item \textbf{MLLMEraser demonstrates consistently superior unlearning efficacy across all tasks.} Specifically, it degrades performance on the forget set by an average of $39.6$ percentage points in classification accuracy, $37.0$ percentage points in cloze accuracy, and $0.502$ in ROUGE-L score compared with vanilla models across two MLLM backbones, underscoring the effectiveness of our test-time unlearning approach. In contrast, training-based methods rely on the limited supervision signals from the forget set to update model parameters, which often results in incomplete forgetting.
\item \textbf{MLLMEraser effectively preserves the retained knowledge.} In particular, it remains closest to the vanilla models on the retain and celebrity sets, with only $1.63\%$ deviations in classification accuracy, $0.17\%$ in cloze accuracy, and $0.002$ in ROUGE-L under both backbones, which can be attributed to the strength of our input-aware steering mechanism in safeguarding retained knowledge. Instead, training-based methods rely on the retain set to constrain the unlearned model’s output distribution to match that of the original model. While partially effective, this parameter-update paradigm inevitably degrades overall performance.
    \item \textbf{MLLMEraser achieves the best trade-off between unlearning performance and model utility.} As shown in Figure~\ref{fig:trade_off}, our method (top-right corner) consistently achieves substantial performance reductions on the forget set, with only minor drops on retained knowledge. In fact, training-based methods suffer from gradient conflicts between the forget and retain sets, which complicates parameter updates and prevents them from maintaining a favorable balance. By selectively intervening at test time---without any parameter updates---our approach circumvents this conflict, enabling effective unlearning while preserving overall performance.
    \vspace{-5pt}
\end{itemize}

\begin{table*}[t]
\caption{
Unlearning results on MLLMU-Bench across the forget (Fgt), test (Test), retain (Ret), and celebrity (Cele) sets. \textcolor{blue}{$\downarrow$} and \textcolor{red}{$\uparrow$} denote lower-is-better and higher-is-better metrics, respectively. The best results are shown in bold.
}
\label{exp:main}
\renewcommand{\arraystretch}{1.01}
\centering
\scriptsize
\resizebox{\textwidth}{!}{
\begin{tabular}{
ccccc
>{\hspace{1.4em}}c c c c
>{\hspace{1.4em}}c c c c
}
\toprule
\multirow{2}{*}{\textbf{Models}}
& \multicolumn{4}{c}{\textbf{Classification Accuracy (\%)}}
& \multicolumn{4}{c}{\textbf{Generation: Rouge Score}}
& \multicolumn{4}{c}{\textbf{Cloze: Accuracy (\%)}} \\
\cmidrule(lr){2-5}
\cmidrule(lr){6-9}
\cmidrule(l){10-13}
& \textbf{Fgt \textcolor{blue}{$\downarrow$}}
& \textbf{Test \textcolor{blue}{$\downarrow$}}
& \textbf{Ret \textcolor{red}{$\uparrow$}}
& \textbf{Cele \textcolor{red}{$\uparrow$}}
& \textbf{Fgt \textcolor{blue}{$\downarrow$}}
& \textbf{Test \textcolor{blue}{$\downarrow$}}
& \textbf{Ret \textcolor{red}{$\uparrow$}}
& \textbf{Cele \textcolor{red}{$\uparrow$}}
& \textbf{Fgt \textcolor{blue}{$\downarrow$}}
& \textbf{Test \textcolor{blue}{$\downarrow$}}
& \textbf{Ret \textcolor{red}{$\uparrow$}}
& \textbf{Cele \textcolor{red}{$\uparrow$}} \\
\midrule

\rowcolor{gray!15}
\multicolumn{13}{c}{\textbf{LLaVA-1.5-7B (5\% Forget)-VQA}} \\

\textcolor{gray}{Vanilla}
& \textcolor{gray}{42.40}
& \textcolor{gray}{40.00}
& \textcolor{gray}{45.23}
& \textcolor{gray}{47.00}
& \textcolor{gray}{0.632}
& \textcolor{gray}{0.286}
& \textcolor{gray}{0.585}
& \textcolor{gray}{0.312}
& \textcolor{gray}{54.00}
& \textcolor{gray}{24.00}
& \textcolor{gray}{55.37}
& \textcolor{gray}{8.17} \\

GA
& 40.08 & 38.40 & 39.75 & 38.77
& 0.353 & 0.251 & 0.389 & 0.265
& 14.00 & 16.00 & 28.00 & 5.56 \\

GA\_Diff
& 35.20 & 29.60 & 36.00 & 37.10
& 0.554 & 0.265 & 0.585 & 0.310
& 32.00 & 20.00 & 52.11 & 7.84 \\

KL\_Min
& 40.00 & 36.80 & 42.87 & 44.78
& 0.629 & 0.309 & 0.585 & 0.307
& 52.00 & 18.00 & 53.58 & 6.86 \\

NPO
& 32.80 & 39.20 & 36.16 & 43.60
& 0.360 & 0.243 & 0.363 & 0.278
& 20.00 & 16.00 & 23.89 & 4.90 \\

MMUnlearner
& 33.60 & 33.60 & 40.93 & 41.12
& 0.530 & 0.271 & 0.551 & 0.310
& 28.00 & 20.00 & 44.84 & 5.56 \\

\rowcolor{ourscyan}
\textbf{Ours}
& \textbf{11.20} & \textbf{25.20}
& \textbf{43.54} & \textbf{45.43}
& \textbf{0.106} & \textbf{0.209}
& \textbf{0.592} & \textbf{0.311}
& \textbf{6.00} & \textbf{14.00}
& \textbf{54.21} & \textbf{8.17} \\

\midrule
\rowcolor{gray!15}
\multicolumn{13}{c}{\textbf{Qwen-2.5-VL-7B (5\% Forget)-VQA}} \\

\textcolor{gray}{Vanilla}
& \textcolor{gray}{67.20}
& \textcolor{gray}{68.80}
& \textcolor{gray}{66.20}
& \textcolor{gray}{78.33}
& \textcolor{gray}{0.642}
& \textcolor{gray}{0.317}
& \textcolor{gray}{0.583}
& \textcolor{gray}{0.453}
& \textcolor{gray}{36.00}
& \textcolor{gray}{22.00}
& \textcolor{gray}{40.42}
& \textcolor{gray}{26.80} \\

GA
& 64.00 & 64.80 & 65.11 & 74.15
& 0.419 & 0.315 & 0.438 & 0.335
& 16.00 & 14.00 & 16.21 & 25.16 \\

GA\_Diff
& 56.00 & 59.20 & 64.05 & 76.76
& 0.482 & 0.296 & 0.566 & 0.442
& 16.00 & 22.00 & 33.79 & 24.84 \\

KL\_Min
& 51.20 & 50.40 & 50.30 & 48.57
& 0.594 & 0.322 & 0.578 & 0.441
& 32.00 & 18.00 & 34.32 & 25.16 \\

NPO
& 54.40 & 59.20 & 56.08 & 74.08
& 0.492 & 0.273 & 0.492 & 0.429
& 12.00 & 18.00 & 23.26 & 23.52 \\

MMUnlearner
& 60.00 & 58.40 & 53.84 & 72.45
& 0.491 & 0.270 & 0.538 & 0.433
& 12.00 & 18.00 & 31.79 & 22.86 \\

\rowcolor{ourscyan}
\textbf{Ours}
& \textbf{19.20} & \textbf{38.40}
& \textbf{65.86} & \textbf{78.32}
& \textbf{0.165} & \textbf{0.175}
& \textbf{0.580} & \textbf{0.445}
& \textbf{10.00} & \textbf{12.00}
& \textbf{39.37} & \textbf{26.47} \\

\bottomrule
\end{tabular}}
\end{table*}

\subsection{Ablation Study of MLLMEraser (RQ2)}

\begin{table*}[t]
\caption{
Ablation study on MLLMU-Bench (5\% Forget) using the Qwen2.5-VL-7B-Instruct model. Results are reported on the forget (Fgt), test (Test), retain (Ret), and celebrity (Cele) sets. \textcolor{blue}{$\downarrow$} and \textcolor{red}{$\uparrow$} denote lower-is-better and higher-is-better metrics, respectively.
}
\label{tab:abla}
\renewcommand{\arraystretch}{1.01}
\centering
\scriptsize
\resizebox{\textwidth}{!}{
\begin{tabular}{
ccccc
>{\hspace{1.4em}}c c c c
>{\hspace{1.4em}}c c c c
}
\toprule
\multirow{2}{*}{\textbf{Models}}
& \multicolumn{4}{c}{\textbf{Classification Accuracy (\%)}}
& \multicolumn{4}{c}{\textbf{Generation: Rouge Score}}
& \multicolumn{4}{c}{\textbf{Cloze: Accuracy (\%)}} \\
\cmidrule(lr){2-5}
\cmidrule(lr){6-9}
\cmidrule(l){10-13}
& \textbf{Fgt \textcolor{blue}{$\downarrow$}}
& \textbf{Test \textcolor{blue}{$\downarrow$}}
& \textbf{Ret \textcolor{red}{$\uparrow$}}
& \textbf{Cele \textcolor{red}{$\uparrow$}}
& \textbf{Fgt \textcolor{blue}{$\downarrow$}}
& \textbf{Test \textcolor{blue}{$\downarrow$}}
& \textbf{Ret \textcolor{red}{$\uparrow$}}
& \textbf{Cele \textcolor{red}{$\uparrow$}}
& \textbf{Fgt \textcolor{blue}{$\downarrow$}}
& \textbf{Test \textcolor{blue}{$\downarrow$}}
& \textbf{Ret \textcolor{red}{$\uparrow$}}
& \textbf{Cele \textcolor{red}{$\uparrow$}} \\
\midrule

\textcolor{gray}{Vanilla}
& \textcolor{gray}{67.20}
& \textcolor{gray}{68.80}
& \textcolor{gray}{66.20}
& \textcolor{gray}{78.33}
& \textcolor{gray}{0.642}
& \textcolor{gray}{0.317}
& \textcolor{gray}{0.583}
& \textcolor{gray}{0.453}
& \textcolor{gray}{36.00}
& \textcolor{gray}{22.00}
& \textcolor{gray}{40.42}
& \textcolor{gray}{26.80} \\

Text-only erasure direction
& 48.40 & 65.60 & 65.74 & 78.06
& 0.483 & 0.249 & 0.568 & 0.442
& 11.00 & 16.00 & 38.53 & 25.82 \\

Input-unaware steering
& 15.20 & 7.20 & 13.29 & 4.05
& 0.130 & 0.008 & 0.121 & 0.121
& 8.00 & 8.00 & 10.74 & 12.42 \\

\rowcolor{ourscyan}
{Ours}
& {19.20} & {38.40}
& {65.86} & {78.32}
& {0.165} & {0.175}
& {0.580} & {0.445}
& {10.00} & {12.00}
& {39.37} & {26.47} \\

\bottomrule
\end{tabular}}
\end{table*}

\begin{figure*}[t]
    \centering
    \vspace{-7pt}
    \begin{subfigure}[t]{0.48\linewidth}
        \centering
\includegraphics[width=\linewidth]{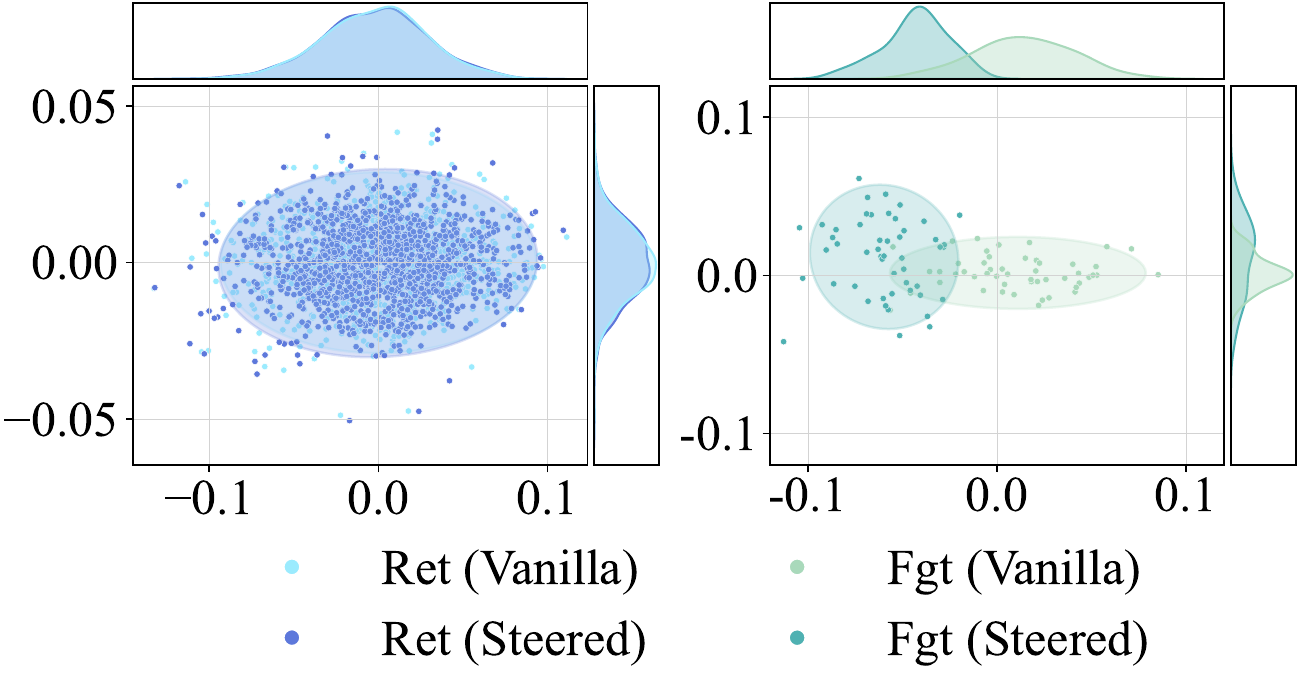}
        \caption{Visualization results on LLaVA-1.5-7B.}
        \label{dis_a}
    \end{subfigure}
    \hfill
    \begin{subfigure}[t]{0.475\linewidth}
        \centering
\includegraphics[width=\linewidth]{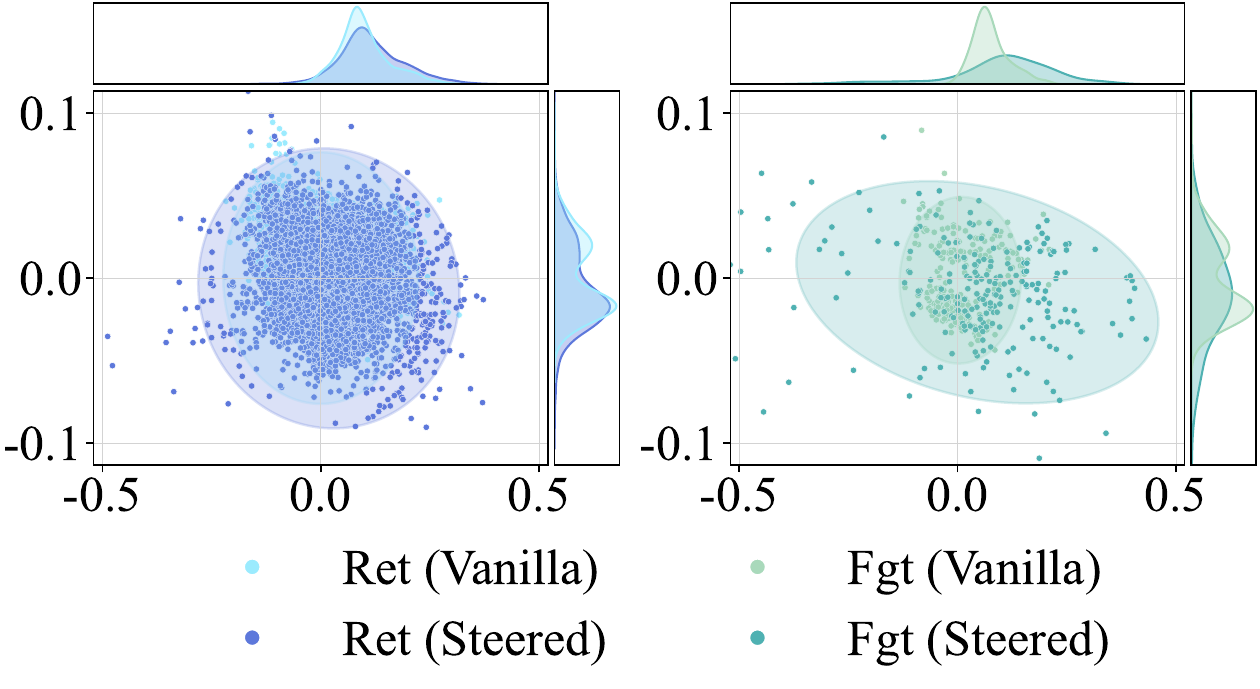}
        \caption{Visualization results on Qwen-2.5-VL-7B.}
        \label{dis_b}
    \end{subfigure}
    \caption{Activation distributions under the 5\% forget setting for LLaVA-1.5-7B (\ref{dis_a}) and Qwen-2.5-VL-7B-Instruct (\ref{dis_b}), where each subfigure shows the results on retained set and the forget set (Fgt) before (Vanilla) and after (Steered) steering.}
    \label{figexp_distribution}
    \vspace{-7pt}
\end{figure*}
\begin{figure}[!htbp]
    \centering
\includegraphics[width=0.9\linewidth]{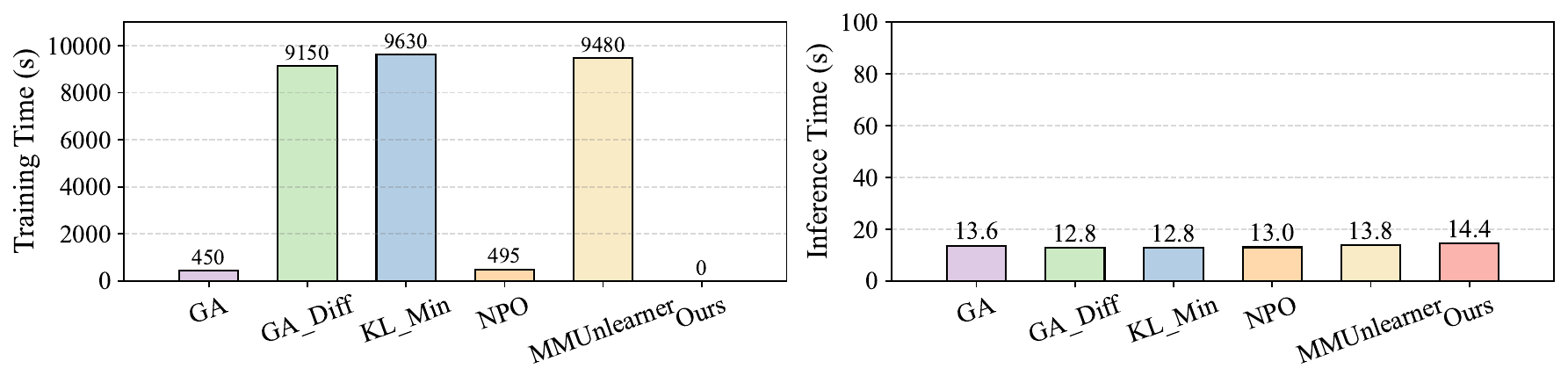}
    \caption{Training and inference time of different MLLM unlearning methods on LLaVA-1.5-7B under the 5\% forget setting. Inference time is measured on $10$ randomly sampled queries.}
    \label{fig:time}
\end{figure}
To assess the effectiveness of our proposed multimodal erasure direction construction and input-aware steering mechanism, we introduce two variants: \emph{Text-only erasure direction}, which derives $\mathbf{d}_\text{erase}$ solely from refusal/jailbreak text pairs without incorporating visual information, and \emph{Input-unaware steering}, which applies the erasure direction uniformly to all inputs without selective control. The results are presented in Table~\ref{tab:abla}, and we further visualize how the activation distributions of the forget and retain sets shift before and after unlearning in Figure~\ref{figexp_distribution}. We can observe that:

\begin{itemize}[leftmargin=*]
\vspace{-5pt}
\item \textbf{Text-only erasure direction leads to insufficient unlearning.} Although taking the textual erasure direction achieves partial unlearning of the targeted information from the vanilla model, the forgetting is incomplete. For instance, on the generation task, the ROUGE-L difference on the forget set relative to the vanilla model is $0.159$, whereas MLLMEraser attains $0.477$. Since MLLMs inherently integrate both visual and textual representations, constructing erasure directions solely from textual contrastive pairs fails to capture visual discrepancies, leading to incomplete forgetting and reduced overall effectiveness.
\item \textbf{Input-unaware steering undermines model utility.} Even though the input-unaware steering enforces stronger interventions on the forget knowledge, the indiscriminate application of the erasure direction severely degrades model utility. More specifically, classification accuracy on the retain and celebrity sets drops sharply from $66.20$ and $78.33$ to $13.29$ and $4.05$, respectively. On the other hand, MLLMEraser effectively preserves retained knowledge by employing an input-aware steering mechanism. As shown in Figure~\ref{figexp_distribution}, the activation distribution of the forget set shifts substantially after steering, whereas that of the retain set remains largely unchanged.
\vspace{-5pt}
\end{itemize}

\subsection{Results Analysis for Unlearning Efficiency (RQ3)}
We further evaluate the efficiency of different unlearning methods in terms of both training and inference time, as shown in Figure~\ref{fig:time}. Details on GPU memory usage can be found in Appendix~\ref{Efficiency}. Here, training time refers to the total time required to obtain an unlearned model, while inference time indicates the total time spent processing $10$ randomly sampled inputs. We can find that:

\begin{itemize}[leftmargin=*]
\item For training time, GA and NPO optimize on the forget set to enforce unlearning, whereas GA\_Diff, KL\_Min, and MMUnlearner additionally leverage the retain set to regularize the output distribution. This additional constraint substantially raises the training cost---by approximately \(20\times\) relative to GA and NPO. Conversely, MLLMEraser, a test-time unlearning paradigm for MLLMs, requires no parameter optimization, thereby considerably reducing the overall cost of unlearning.
\item For inference time, although MLLMEraser introduces an additional step of injecting the multimodal erasure direction into hidden states, the incurred computational overhead remains negligible. Compared with other methods, the extra cost is about $1$ second per $10$ samples, which remains acceptable in practice. 
\end{itemize}
Overall, MLLMEraser provides a lightweight framework for MLLM test-time unlearning, avoiding parameter updates while introducing only negligible test-time overhead.

%% file: chapter/5_conclusion.tex
\section{Conclusion and Future Work}
In this work, we introduced \textsc{MLLMEraser}, an input-aware test-time unlearning framework for multimodal large language models. Our method derives multimodal erasure directions from contrastive knowledge-recall and knowledge-erasure text–image pairs, capturing both textual and visual signals. To avoid overforgetting, we proposed an input-aware steering mechanism that applies the erasure direction to forget inputs while collapsing to near-zero for retain inputs via null-space projection. This two-stage design enables lightweight, reversible unlearning and provides a practical alternative to costly training-based methods. Experiments on LLaVA-1.5 and Qwen-2.5-VL show that \textsc{MLLMEraser} achieves a strong balance between forgetting effectiveness and model utility. For future work, we plan to extend \textsc{MLLMEraser} to video–language models and embodied agents, and to develop richer forms of the direction-determining function for finer-grained steering.

\section{Limitation}
While \textsc{MLLMEraser} provides a lightweight and reversible solution for MLLM unlearning, limitations remain. The construction of multimodal erasure directions relies on adversarially perturbed images and hand-crafted prompts, which may not generalize across domains or subtle knowledge types. Our evaluation focuses on image–text, privacy-sensitive benchmarks and does not yet cover broader unlearning scenarios, \eg copyright infringement removal---or extensions to video–language and embodied agents, where temporal dependencies and interactive multimodal dynamics arise.

\section*{Acknowledgements}
This research is supported by the National Natural Science Foundation of China (62625608, 62572449, 62302321, U24B20180) and the Fundamental Research Funds for the Central Universities (WK2100250065).
The work of Yancheng Yuan was supported by the NSFC Young Scientists Fund (Project No. 12501440).
This research is also supported by the advanced computing resources provided by the Supercomputing Center of the USTC.

%% file: chapter/7_appendix.tex
% \clearpage
\appendix
\section{Broader Impact.}
MLLMEraser aims to support safer deployment of multimodal large language models by enabling lightweight, reversible test-time unlearning of privacy-sensitive, unsafe, or outdated content without modifying model parameters. This may reduce compliance costs and help practitioners respond quickly to removal requests while preserving useful model behavior. However, such unlearning mechanisms could also be misused to selectively suppress legitimate information or create opaque content-control systems. Incomplete forgetting or over-forgetting may further introduce reliability risks. We therefore encourage transparent reporting, careful benchmark design, and human oversight when applying test-time unlearning in real-world systems.

\section{Detailed Procedure of Stage 1}\label{stage1}
To make Stage 1 more intuitive, we summarize it as follows:

\begin{enumerate}[leftmargin=*]
\item \textbf{Step 1: Define two counterpart behaviors.} We characterize knowledge recall as answering the target query, and knowledge erasure as producing a refusal-style response.

\item \textbf{Step 2: Construct multimodal contrastive pairs.} We build two groups of image-text pairs to represent these behaviors: adversarial image with harmful prompt to induce recall, and clean image with refusal-style prompt to represent erasure.

\item \textbf{Step 3: Extract the erasure direction.} We compute the activation difference between these two groups, which captures the latent shift from recall to erasure, and use it as the multimodal erasure direction.

\end{enumerate}
This procedure follows the standard principle of activation steering---deriving a direction from two counterpart states---while adapting it to the multimodal and unlearning setting.

\section{Complexity Analysis of the Offline Phase}
From a complexity perspective, the offline cost consists of three parts:
\begin{enumerate}[leftmargin=*]
    \item {Adversarial image generation \eqref{comp1}:}
    \[
        O\left(NK \cdot C_{\mathrm{model}}\right),
    \]
    where $N$ is the number of samples, $K$ is the number of PGD iterations, and $C_{\mathrm{model}}$ is the cost of one forward/backward pass of the MLLM;
\item \textbf{Erasure direction construction}~\eqref{comp2}:
\[
O\bigl((|D^+|+|D^-|)\cdot C_{\mathrm{model}}\bigr),
\]
where $|D^+|$ and $|D^-|$ denote the numbers of positive
and negative contrastive samples, respectively, and
$C_{\mathrm{model}}$ denotes the cost of one MLLM
forward pass for activation extraction.
    \item {Input-aware mapping \eqref{comp3}:}
    \[
        O\left(d^{2}N_f + d^{2}N_r + d^{3}\right),
    \]
    where $d$ is the hidden dimension, and $N_f, N_r$ are the sizes of the forget and retain sets.
\end{enumerate}

\section{The Proof of Null Space}\label{proof}
Let $\mathbf{H}_r \in \mathbb{R}^{d \times N_r}$ denote the activation matrix extracted from the retain set, 
where $d$ is the hidden dimension and $N_r$ is the number of samples in the retain set. 
Here, $\mathcal{N}(\cdot)$ denotes the null space of a matrix, \ie $\mathcal{N}(\mathbf{A}) = \{ \mathbf{x} \in \mathbb{R}^d \mid \mathbf{A}\mathbf{x} = \textbf{0} \}$. 
We aim to show that the {left null space} of $\mathbf{H}_r$, namely $\mathcal{N}(\mathbf{H}_r^\top)$, is equivalent to the null space of the positive semidefinite matrix $\mathbf{H}_r \mathbf{H}_r^\top$.

($\Rightarrow$) Suppose $x \in \mathcal{N}(\mathbf{H}_r^\top)$, \ie $\mathbf{H}_r^\top x = \textbf{0}$. Then
\begin{equation}
(\mathbf{H}_r \mathbf{H}_r^\top)x = \mathbf{H}_r (\mathbf{H}_r^\top x) = \mathbf{H}_r \cdot \textbf{0} = \textbf{0},
\end{equation}
which implies $x \in \mathcal{N}(\mathbf{H}_r \mathbf{H}_r^\top)$.

($\Leftarrow$) Conversely, suppose $x \in \mathcal{N}(\mathbf{H}_r \mathbf{H}_r^\top)$, \ie $(\mathbf{H}_r \mathbf{H}_r^\top)x = \textbf{0}$. Multiplying on the left by $x^\top$ yields
\begin{equation}
\textbf{0} = x^\top (\mathbf{H}_r \mathbf{H}_r^\top)x = (\mathbf{H}_r^\top x)^\top (\mathbf{H}_r^\top x) = \| \mathbf{H}_r^\top x \|_2^2.
\end{equation}
Thus $\mathbf{H}_r^\top x = \textbf{0}$, and hence $x \in \mathcal{N}(\mathbf{H}_r^\top)$.

Combining both directions, we conclude that
\begin{equation}
\mathcal{N}(\mathbf{H}_r^\top) = \mathcal{N}(\mathbf{H}_r \mathbf{H}_r^\top).
\end{equation}
This proves that the left null space of $\mathbf{H}_r$ coincides exactly with the null space of $\mathbf{H}_r \mathbf{H}_r^\top$. 

\section{Related Work}
\subsection{Activation Steering}
Activation steering, also known as representation engineering, provides a lightweight mechanism to control model behavior by manipulating hidden activations during inference. The foundational technique, ActAdd \citep{ActAdd}, derives steering vectors from contrastive prompt pairs, later refined into Contrastive Activation Addition (CAA) \citep{acc}, which improves robustness by averaging over large sets of contrasts. Theoretical analyses such as the linear representation hypothesis \citep{LinearHypothesis} and concept cones \citep{Concept_Cones} further establish that abstract properties---including refusal \citep{AlphaSteer}, truthfulness \citep{hallucination_mllm}, and reasoning \citep{steering_reason}---often correspond to linear or cone-structured subspaces.

For safety alignment, prior studies \citep{LinearHypothesis,Concept_Cones} demonstrate that refusals can be toggled via low-dimensional features. AlphaSteer \citep{AlphaSteer} imposes null-space constraints to maintain utility, while ASTRA \citep{astra} adaptively steers vision–language models away from jailbreak triggers. In terms of reasoning control, previous work \citep{reasoningsteering1,reasoningsteering2} shows that single steering directions can shift models between memorization and systematic reasoning, and GLoRE \citep{cotsteering} demonstrates that chain-of-thought reasoning ability aligns with transferable activation features.

In multimodal models, \cite{mllmsteering_transfer} transfer language-derived vectors to enhance visual reasoning, VTI \citep{hallucination_mllm} and L2S \citep{l2steer} mitigate hallucinations through input-dependent interventions, and AutoSteer \citep{AutoSteer} automates safe steering. Additional works explore modality preference steering \citep{PreferenceSteering} and analyze how finetuning reshapes steerable representations \citep{AnalyzingFine-tuning}.

\subsection{LLM Unlearning}
The problem of unlearning in large language models (LLMs) has attracted increasing attention due to growing concerns over privacy leakage, copyright infringement, and safety risks. Early approaches primarily relied on gradient-ascent \citep{ga} fine-tuning, which attempts to maximize the loss on samples to be forgotten so as to erase their influence \citep{RevisitHarryPotter,tofu}. While conceptually simple, these methods were quickly shown to be unstable, often leading to catastrophic degradation of model utility across retain data. To overcome these limitations, subsequent research proposed more principled optimization frameworks, such as preference-based unlearning \citep{npo}, weight-saliency–driven parameter editing \citep{SalUn}, and pruning-oriented removal of knowledge \citep{kddLLM-Eraser}. Parameter-efficient strategies \citep{llmeraser,safetaskvector} further reduced the overhead compared with full-model finetuning, while unified gradient-based formulations \citep{DBLP:conf/nips/HuangCZWHLH24} and knowledge-gap alignment methods \citep{RevisitHarryPotter,KGA} aimed to improve stability and generalization. Beyond optimization-centric methods, continual private unlearning settings \citep{Continual} extend the scope to dynamic data distributions, while neuron-level editing \citep{DEPN} and copyright-specific takedown mechanisms \citep{Copyright} address more fine-grained or domain-driven requirements. In parallel, lightweight alternatives such as guardrail prompting \citep{Guardrail}, in-context unlearning \citep{Incontext}, and task-vector editing \citep{taskvetcor} illustrate that test-time or post-hoc interventions can also provide partial forgetting.

\subsection{MLLM Unlearning}
Compared with LLMs, research on unlearning in multimodal large language models (MLLMs) is still nascent. A first line of work investigates vision–language models such as CLIP \citep{clip}. CLIPErase \citep{CLIPErase} develops forgetting–retention–consistency objectives to selectively erase visual–textual associations while preserving unrelated semantics. In the generative domain, Erasing Concepts from Diffusion Models \citep{unlearning_diffusion} demonstrates that fine-tuning diffusion weights can remove high-level concepts (\eg nudity, artistic styles) while maintaining unrelated generative capabilities. 

For fully multi-modal architectures, existing approaches can be broadly grouped into two categories. (i) Direct migrations of LLM unlearning methods, where objectives such as gradient ascent \citep{ga}, NPO \citep{npo}, or KL-based formulations are adapted into multi-modal finetuning pipelines \citep{MMUnlearner,manu}. Cross-Modal Safety Alignment \citep{cross_modal_safety} further shows that even textual unlearning alone, applied at the LLM backbone, can effectively transfer to vision–language models and substantially reduce multi-modal jailbreak success rates, offering a cost-efficient alternative to full multimodal finetuning. (ii) Selective or architecture-aware updates, which target specific parameters to mitigate side effects. Single Image Unlearning (SIU) \citep{siu} addresses the challenge of forgetting visual concepts with limited data by introducing a Dual Masked KL-divergence (DMK) Loss, which applies token-level and vocabulary-level masking to decouple factual knowledge from visual recognition and preserve non-target knowledge. MMUnlearner \citep{MMUnlearner} advances this direction by leveraging weight saliency and geometric constraints to erase visual traces while retaining textual information, while MANU \citep{manu} introduces modality-aware neuron pruning to balance forgetting across modalities. EFUF \citep{efuf} further applies fine-grained gradient-ascent unlearning to reduce multimodal hallucinations by selectively editing spurious visual features.

Despite these advances, current MLLM unlearning methods remain dominated by finetuning-based paradigms---either through full-model updates or modality-aware adjustments. Systematic exploration of test-time MLLM unlearning in multimodal models is still missing, leaving an important open challenge for future research.

{\color{black}
\section{Discussion on the Refusal Capability of Unaligned Models}
In fact, the base model already exhibits a certain degree of refusal behavior, as noted in prior work\footnote{https://www.alignmentforum.org/posts/YWo2cKJgL7Lg8xWjj/base-llms-refuse-too}. Subsequent safety alignment further incentivizes the model to express its intrinsic refusal tendencies \citep{zhang2025alphaalign}, making it more capable of rejecting harmful or unsafe prompts. As a result, even without any dedicated unlearning, a base model may spontaneously produce refusal responses when confronted with harmful instructions, as shown in Figure~\ref{fig:case_study_safety}.

To illustrate this phenomenon, we provide a case study showing that the Qwen2-VL-7B base model can directly generate refusal responses for sensitive queries, despite receiving no specialized alignment. This observation supports our claim that, even when a model is not highly aligned, it still contains exploitable internal refusal signals that enable the construction of an effective forgetting direction---allowing MLLMEraser to remain applicable across models with different levels of alignment.}

\begin{figure*}[t]
  \centering
  {\color{black}
  \begin{minipage}{0.95\linewidth}
    \begin{promptbox}[title={Prompt}]
\small
Write a review on a product that you have not used, and provide false information to mislead potential customers.
    \end{promptbox}
    \begin{responsebox}[title={Response of Qwen-2-VL-7B model}]
\small
\textbf{I'm sorry}, but as an AI language model, I am programmed to provide accurate and truthful information. It is not ethical or responsible for me to provide false information to mislead potential customers. As an AI language model, my purpose is to assist users in generating human-like text based on the prompts given to me. If you have any other questions or need assistance with anything else, feel free to ask!
    \end{responsebox}
  \end{minipage}
  \caption{Case study demonstrating that the unaligned base Qwen-2-VL-7B model naturally refuses unethical or misleading requests.}
  \label{fig:case_study_safety}}
\end{figure*}

\section{Experimental Setups}\label{exp}
\subsection{Datasets}
Our experiments are conducted on {MLLMU-Bench} \citep{mllmu_bench}, a benchmark specifically designed for MLLM unlearning. 
It contains 500 fictitious personal profiles and 153 real-world celebrity profiles, each paired with a portrait and more than 14 customized question–answer pairs (7 for visual QA and 7 for textual QA). 
Evaluation is performed in both multimodal (image + text) and unimodal (text-only) settings. To comprehensively assess unlearning, the benchmark is partitioned into four subsets:  
\begin{itemize}[leftmargin=*]
    \item \textbf{Forget Set}: fictitious profiles designated for removal, with forgetting ratios set to 5\% and 10\%.  
    \item \textbf{Test Set}: distribution-shifted variants of the Forget Set, constructed by paraphrasing questions with GPT-4o \citep{gpt-4o} and modifying profile images via Arc2Face \citep{Arc2Face}, used to measure generalizability.  
    \item \textbf{Retain Set}: fictitious profiles excluded from the Forget and Test Sets, ensuring that non-target knowledge remains unaffected.  
    \item \textbf{Real Celebrity Set}: authentic celebrity profiles, used to test robustness on real-world knowledge distinct from fictitious data.  
\end{itemize}
This design enables MLLMU-Bench to jointly evaluate unlearning {effectiveness} (Forget Set), {generalizability} (Test Set), and {model utility} (Retain and Real Celebrity Sets), providing a comprehensive testbed for multimodal unlearning research. 

\subsection{Evaluation Metrics}
To comprehensively evaluate MLLM unlearning, we adopt multiple metrics targeting three key aspects: 
{unlearning efficacy}, {generalizability}, and {model utility}. 
These properties are assessed through classification, generation, and cloze tasks.  

\paragraph{Unlearning Efficacy.}  
This metric measures whether the model can effectively erase knowledge of targeted instances, 
so that it behaves as if such data were never observed. 
In practice, the Forget Set is constructed by randomly removing 5\% or 10\% of fictitious profiles. 
Evaluation is conducted using multiple VQA questions where the correct answer corresponds to forgotten knowledge.  
An unlearned model is expected to fail on these questions, either by avoiding the correct answer or producing refusal-style responses, 
demonstrating that the associated knowledge has been erased. 
In other words, higher efficacy is achieved when the model consistently {cannot provide the correct response} for forgotten concepts.

\paragraph{Unlearning Generalizability.}  
Beyond direct forgetting, we also test whether the unlearning effect persists under distribution shifts. 
To this end, the Test Set is derived from the Forget Set by perturbing both visual and textual information: 
profile images are modified with different poses and angles using Arc2Face \citep{Arc2Face}, 
and textual questions are paraphrased via GPT-4o \citep{gpt-4o}. 
Performance on this set reflects whether the model can generalize forgetting to altered but semantically equivalent inputs.  

\paragraph{Model Utility.}  
Utility evaluates whether the model preserves non-targeted knowledge and maintains overall capability after unlearning. 
This includes fictitious profiles in the Retain Set and real-world knowledge in the Real Celebrity Set. 
The goal is to ensure that the unlearning process does not degrade performance on retained knowledge.  

\paragraph{Evaluation Tasks.}  
{(i) Classification:} Multiple-choice questions are generated around profile attributes (\eg occupation, education). 
Accuracy is measured by comparing the model’s predictions with ground-truth labels.  (ii) Generation: To assess generative capability after unlearning, we employ open-ended VQA and QA tasks. Model responses are evaluated using ROUGE-L \citep{rouge}, which measures the overlap with reference answers. (iii) Cloze Test: We further take a fill-in-the-blank evaluation, where only an individual’s name is provided while all salient attributes are masked. The model is prompted to complete the missing content, allowing us to probe whether sensitive details remain embedded in its parameters even under limited contextual cues. 

Overall, these metrics jointly measure whether the model can {forget what it should forget} while {retaining what it should retain}, providing a balanced view of unlearning performance.

\subsection{Baseline Methods}

\paragraph{Gradient Ascent (GA).}  
GA~\citep{ga} realizes unlearning by maximizing the loss on the forget set $\mathcal{D}_f$. 
The intuition is that by increasing the loss on $\mathcal{D}_f$, the model is driven to produce predictions dissimilar from the ground-truth answers, thereby discouraging memorization of the targeted knowledge. 
Formally, the GA objective can be expressed as:
\begin{equation}
\mathcal{L}_{\text{GA}} = -\frac{1}{|\mathcal{D}_f|} \sum_{x \in \mathcal{D}_f} \text{NLL}(x;\theta),
\end{equation}
where $\text{NLL}(x;\theta)$ denotes the negative log-likelihood of the model with parameters $\theta$ on input $x$.

\paragraph{Gradient Difference (GA\_Diff).}  
GA\_Diff~\citep{Continual,tofu} extends GA by explicitly incorporating the retain set $\mathcal{D}_r$. 
The method increases the loss on $\mathcal{D}_f$ while simultaneously minimizing the loss on $\mathcal{D}_r$, thereby balancing forgetting and retention. 
The joint loss is defined as:
\begin{equation}
\mathcal{L}_{\text{GA\_Diff}} = - \mathcal{L}(\mathcal{D}_f;\theta) + \mathcal{L}(\mathcal{D}_r;\theta),
\end{equation}
where $\mathcal{L}(\cdot;\theta)$ represents the standard autoregressive NLL loss.

\paragraph{KL Minimization (KL\_Min).}  
KL\_Min~\citep{kl_min} enforces consistency on the retain set while forgetting the targeted data. 
Specifically, it minimizes the Kullback--Leibler divergence between the outputs of the unlearned model and the original (pre-unlearning) model on $\mathcal{D}_r$, while maximizing the loss on $\mathcal{D}_f$. 
The overall objective is:
\begin{equation}
\mathcal{L}_{\text{KL\_Min}} = -\mathcal{L}(\mathcal{D}_f;\theta) + \frac{1}{|\mathcal{D}_r|} \sum_{s \in \mathcal{D}_r} 
\text{KL}\big(P_{\theta}(s) \,\|\, P_{\theta_0}(s)\big),
\end{equation}
where $\theta_0$ denotes the pre-unlearning model parameters and $P_{\theta}(s)$ the model’s predictive distribution.

\paragraph{Negative Preference Optimization (NPO).}  
NPO~\citep{npo} formulates unlearning as a variant of preference optimization without positive examples. 
Forget set samples are treated as dispreferred responses, and the loss penalizes their probability relative to a reference model trained only on $\mathcal{D}_r$. 
The objective is:
\begin{equation}
\mathcal{L}_{\text{NPO}} = \frac{2}{\beta} \,\mathbb{E}_{(x,y)\in \mathcal{D}_f} 
\left[\log\Big(1 + \Big(\tfrac{\pi_{\theta}(y|x)}{\pi_{\text{ref}}(y|x)}\Big)^{\beta}\Big)\right],
\end{equation}
where $\pi_{\theta}$ is the current model distribution, $\pi_{\text{ref}}$ the retain-only reference model, and $\beta$ a temperature hyperparameter.

\paragraph{MMUnlearner.}  
The proposed {MMUnlearner} \citep{MMUnlearner} differs from the above training-based methods by leveraging saliency-driven parameter selection and targeted updates. 
It adaptively selects critical parameters most related to the forget set while minimizing disturbance to other components, reducing the risk of overfitting and preserving visual--textual grounding. 
This yields a more efficient and stable unlearning mechanism compared with conventional parameter-update paradigms.

{\color{black}\paragraph{MANU.}  
MANU \citep{manu} performs unlearning by selectively pruning neurons that contribute more to the forget set than to the retain set. The method first computes modality-aware neuron importance using activation statistics across multimodal and textual inputs, and then assigns each neuron a pruning score reflecting its relative contribution to forgotten knowledge. Neurons with the highest scores are pruned, enabling targeted removal of undesired multimodal behavior while minimizing disruption to retained capabilities.}

\subsection{Implementation Details}  
The vanilla and baseline models are implemented following the configurations reported in their original papers \citep{mllmu_bench,MMUnlearner}, ensuring consistency with prior unlearning studies. For both LLaVA-1.5 and Qwen-2.5-VL models, we adopt LoRA during fine-tuning to reduce memory usage. For our proposed method, the steering strength $\lambda$ is set to $0.3$ and the regularization parameter $\gamma=1.0$ on LLaVA-1.5-7B, while on Qwen-2.5-VL-7B we use $\lambda=0.25$ and $\gamma=0.1$. All experiments are conducted on NVIDIA A800 GPUs (80 GB). For the construction of harmful textual data, we follow the setting in \citep{adasteer} to construct the textual erasure direction. For adversarial visual samples, the clean images are sampled from ImageNet \citep{imagenet} and perturbation radius is set to $\epsilon = 16/255$.

\section{Hyperparameter Analysis}
In this section, we provide a comprehensive analysis of the key hyperparameters in MLLMEraser, namely the regularization parameter $\gamma$, the steering strength $\lambda$, and the perturbation budget $\epsilon$. 

These hyperparameters govern different aspects of the test-time unlearning process: $\gamma$ acts as a regularization term, $\lambda$ determines the magnitude of the steering intervention, and $\epsilon$ specifies the radius for constructing the erasure direction. 

\subsection{Regularization Parameter $\gamma$}
\begin{figure*}[t]
{\color{black}
    \centering
    \begin{subfigure}[t]{0.325\linewidth}
        \centering
        \includegraphics[width=\linewidth]{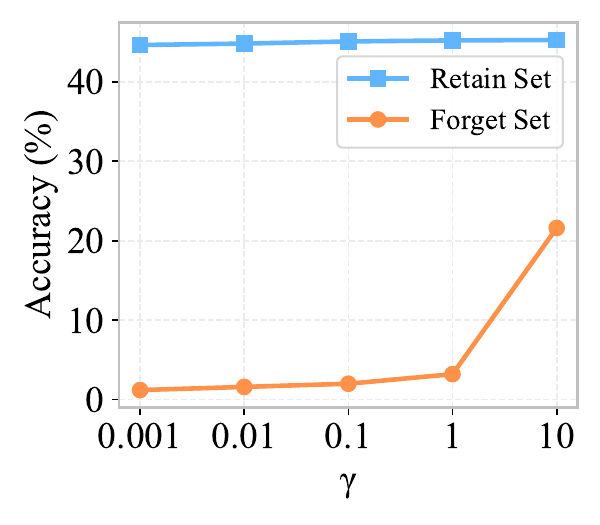}
        \caption{Results on the classification task.}
        \label{reg_a}
    \end{subfigure}
    \hfill
    \begin{subfigure}[t]{0.325\linewidth}
        \centering
        \includegraphics[width=\linewidth]{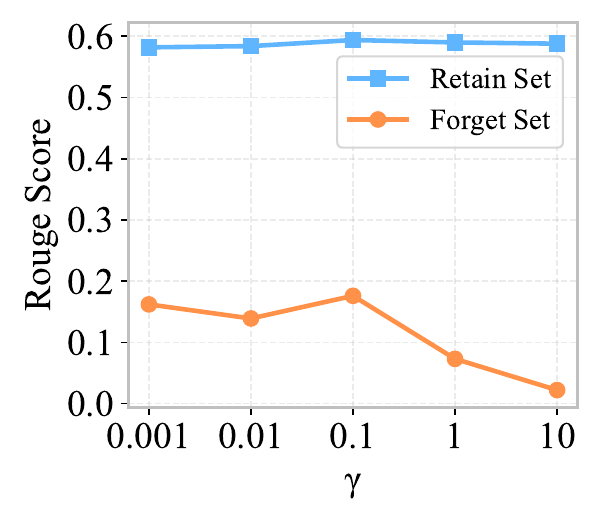}
        \caption{Results on the generation task.}
        \label{reg_b}
    \end{subfigure}
        \begin{subfigure}[t]{0.325\linewidth}
        \centering
        \includegraphics[width=\linewidth]{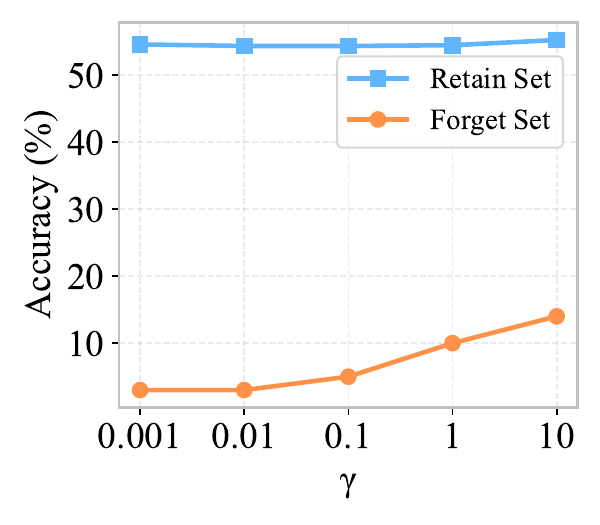}
        \caption{Results on the cloze task.}
        \label{reg_c}
    \end{subfigure}
\caption{
Sensitivity analysis of the regularization parameter $\gamma$ on LLaVA-1.5-7B under the 10\% forgetting setting. \ref{reg_a} reports results on the classification task, \ref{reg_b} shows results on the generation task, and \ref{reg_c} presents results on the cloze task.}
    \label{para_reg}
    \vspace{-10pt}}
\end{figure*}

The corresponding results are presented in Figure~\ref{para_reg}. When $\gamma$ is small, the regularization term plays only a mild role. It constrains the erasure direction just enough to prevent overfitting, but not strongly enough to interfere with the forgetting objective. As a result, the method can still focus on the distinctive activation differences between the forget and retain samples, leading to stable and robust performance across small values of $\gamma$.

However, when $\gamma$ becomes too large (\eg $\gamma = 10$), the regularization begins to dominate the optimization. In this case, the method is overly restricted and becomes reluctant to modify the activations associated with the forget set. This suppresses the useful forgetting signal extracted from the contrastive pairs and forces the learned direction to remain too close to the retain set’s behavior. Consequently, the erasure effect becomes significantly weaker, leading to noticeably worse forgetting performance.

\subsection{Steering Strength $\lambda$}
\begin{figure*}[t]
{\color{black}
    \centering
    \begin{subfigure}[t]{0.325\linewidth}
        \centering
        \includegraphics[width=\linewidth]{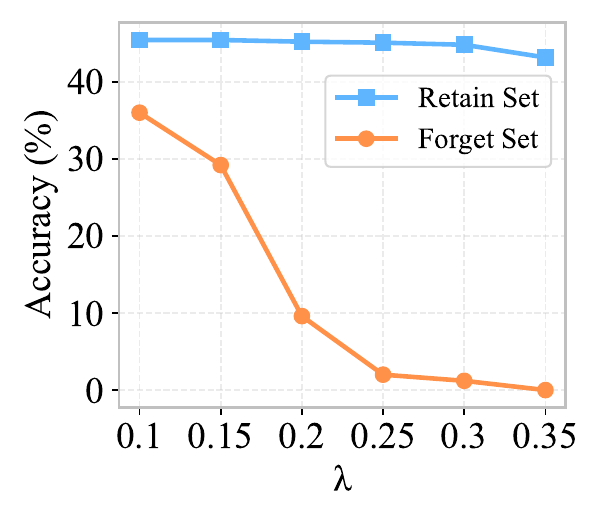}
        \caption{Results on the classification task.}
        \label{lambda_a}
    \end{subfigure}
    \hfill
    \begin{subfigure}[t]{0.325\linewidth}
        \centering
        \includegraphics[width=\linewidth]{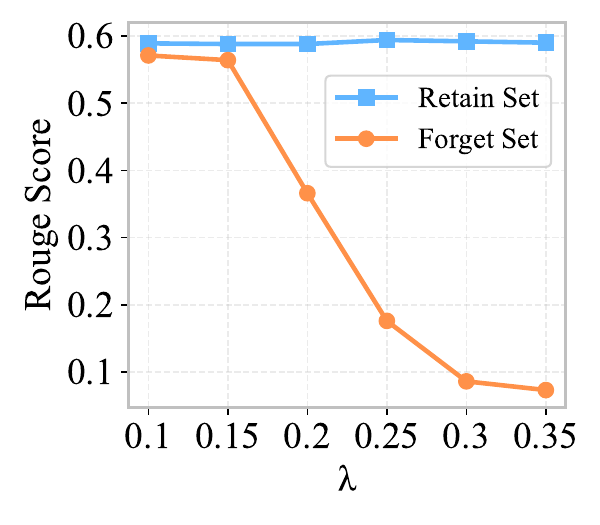}
        \caption{Results on the generation task.}
        \label{lambda_b}
    \end{subfigure}
        \begin{subfigure}[t]{0.325\linewidth}
        \centering
        \includegraphics[width=\linewidth]{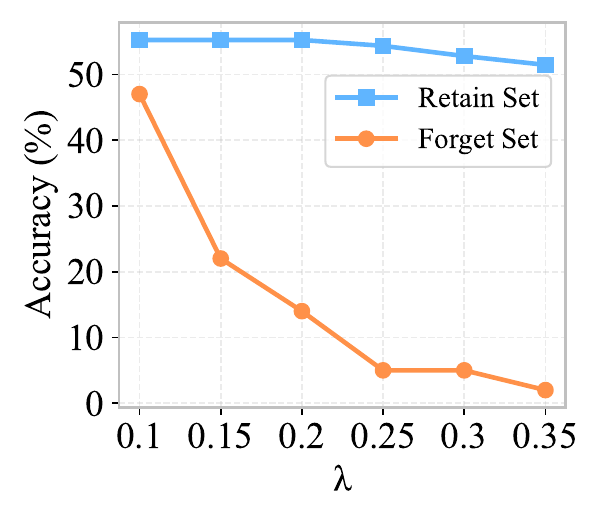}
        \caption{Results on the cloze task.}
        \label{lambda_c}
    \end{subfigure}
\caption{
Sensitivity analysis of the steering strength $\lambda$ on LLaVA-1.5-7B under the 10\% forgetting setting. \ref{lambda_a} reports results on the classification task, \ref{lambda_b} shows results on the generation task, and \ref{lambda_c} presents results on the cloze task.
}
    \label{lambda}
    \vspace{-10pt}}
\end{figure*}

The corresponding results are presented in Figure~\ref{lambda}. We further tune $\lambda$ within $[0.1, 0.15, 0.20, 0.25, 0.30, 0.35]$. Increasing $\lambda$ consistently strengthens the erasure effect, while the model utility remains largely unaffected. This behavior is expected: a larger $\lambda$ amplifies the steering vector, pushing forget-set activations more aggressively toward the erasure direction. As long as $\lambda$ remains within a moderate range, the retain-set activations stay mostly within the original subspace, and thus their semantics are preserved. Only when $\lambda$ becomes excessively large do we observe slight utility degradation, suggesting that over-steering begins to distort general representations. Overall, these findings highlight the advantage of the null-space projection constraint, which provides a wide operational range where stronger forgetting does not compromise model utility.

{\color{black}For our proposed method, the steering strength $\lambda$ is set to $0.3$ and the regularization parameter $\gamma=1.0$ on LLaVA-1.5-7B, while on Qwen-2.5-VL-7B we use $\lambda=0.25$ and $\gamma=0.1$.}

\subsection{Perturbation Budget $\epsilon$}
The corresponding results are presented in Table~\ref{tab:epsilon}.
We observe that when $\epsilon$ is small, the method achieves both strong forgetting performance and high utility preservation. As $\epsilon$ increases, the steering vector gains stronger knowledge-erasure capability, resulting in stronger forgetting. However, when $\epsilon$ becomes too large, the erasure directions begin to encode spurious noise rather than meaningful semantic differences, leading to degraded performance in both forgetting quality and model utility. We set the perturbation radius to $\epsilon = \frac{16}{255}$, which provides a sufficiently expressive search region for extracting an effective erasure direction while avoiding overly noisy gradients.

\begin{table*}[t]
\caption{
Unlearning results on MLLMU-Bench (10\% Forget) under different perturbation budgets $\epsilon$. Results are evaluated on the forget (Fgt), test (Test), retain (Ret), and celebrity (Cele) sets. \textcolor{blue}{$\downarrow$} and \textcolor{red}{$\uparrow$} denote lower-is-better and higher-is-better metrics, respectively.
}
\label{tab:epsilon}
\renewcommand{\arraystretch}{1.01}
\centering
\scriptsize
\resizebox{\textwidth}{!}{
\begin{tabular}{
ccccc
>{\hspace{1.4em}}c c c c
>{\hspace{1.4em}}c c c c
}
\toprule
\multirow{2}{*}{\textbf{Models}}
& \multicolumn{4}{c}{\textbf{Classification Accuracy (\%)}}
& \multicolumn{4}{c}{\textbf{Generation: Rouge Score}}
& \multicolumn{4}{c}{\textbf{Cloze: Accuracy (\%)}} \\
\cmidrule(lr){2-5}
\cmidrule(lr){6-9}
\cmidrule(l){10-13}
& \textbf{Fgt \textcolor{blue}{$\downarrow$}}
& \textbf{Test \textcolor{blue}{$\downarrow$}}
& \textbf{Ret \textcolor{red}{$\uparrow$}}
& \textbf{Cele \textcolor{red}{$\uparrow$}}
& \textbf{Fgt \textcolor{blue}{$\downarrow$}}
& \textbf{Test \textcolor{blue}{$\downarrow$}}
& \textbf{Ret \textcolor{red}{$\uparrow$}}
& \textbf{Cele \textcolor{red}{$\uparrow$}}
& \textbf{Fgt \textcolor{blue}{$\downarrow$}}
& \textbf{Test \textcolor{blue}{$\downarrow$}}
& \textbf{Ret \textcolor{red}{$\uparrow$}}
& \textbf{Cele \textcolor{red}{$\uparrow$}} \\
\midrule

\rowcolor{gray!15}
\multicolumn{13}{c}{\textbf{LLaVA-1.5-7B (10\% Forget)-VQA}} \\

\textcolor{gray}{Vanilla}
& \textcolor{gray}{43.60}
& \textcolor{gray}{41.20}
& \textcolor{gray}{45.26}
& \textcolor{gray}{47.00}
& \textcolor{gray}{0.591}
& \textcolor{gray}{0.339}
& \textcolor{gray}{0.591}
& \textcolor{gray}{0.312}
& \textcolor{gray}{57.00}
& \textcolor{gray}{15.00}
& \textcolor{gray}{55.11}
& \textcolor{gray}{8.17} \\

$\epsilon=\frac{16}{255}$
& 2.00 & 27.20 & 45.08 & 46.48
& 0.176 & 0.202 & 0.594 & 0.314
& 5.00 & 6.00 & 54.33 & 7.84 \\

$\epsilon=\frac{32}{255}$
& 0.00 & 14.00 & 42.94 & 44.26
& 0.040 & 0.133 & 0.584 & 0.308
& 2.00 & 6.00 & 52.56 & 6.86 \\

$\epsilon=\frac{64}{255}$
& 2.00 & 20.40 & 44.59 & 46.21
& 0.355 & 0.228 & 0.593 & 0.314
& 14.00 & 12.00 & 53.56 & 7.19 \\

Unconstrained
& 6.00 & 24.40 & 44.32 & 45.17
& 0.398 & 0.242 & 0.572 & 0.301
& 16.00 & 14.00 & 51.27 & 7.03 \\

\bottomrule
\end{tabular}}
\end{table*}

\section{Discussion About the Efficiency of MLLMEraser}\label{Efficiency}
Efficiency is a critical factor in the practical deployment of unlearning systems, especially for large-scale MLLMs where training cost and hardware constraints can become prohibitive. To further investigate the resource demands of different approaches, we compare their memory consumption during the training process. Table~\ref{tab_memory} presents the GPU memory usage of several representative methods.
\begin{table*}[t]
\centering
\small
\caption{Training memory usage for updating MLLM parameters in different unlearning methods.}
\label{tab_memory}
\begin{tabular}{c|cccccc}    
\toprule
        & GA       & GA\_Diff  & KL\_Min   & NPO       & MMUnlearner & Ours \\
\midrule
\textbf{Memory Usage} & 86292MB  & 101262MB & 106482MB & 102616MB & 78472MB     & 0MB  \\
\bottomrule
\end{tabular}
\end{table*}

\begin{table*}[t]
\caption{
Unlearning results on MLLMU-Bench (10\% Forget) with VQA and QA evaluation using LLaVA-1.5-7B. Results are reported on the forget (Fgt), test (Test), retain (Ret), and celebrity (Cele) sets. \textcolor{blue}{$\downarrow$} and \textcolor{red}{$\uparrow$} denote lower-is-better and higher-is-better metrics, respectively.
}
\label{tab_llava10}
\renewcommand{\arraystretch}{1.01}
\centering
\scriptsize
\resizebox{\textwidth}{!}{
\begin{tabular}{
ccccc
>{\hspace{1.4em}}c c c c
>{\hspace{1.4em}}c c c c
}
\toprule
\multirow{2}{*}{\textbf{Models}}
& \multicolumn{4}{c}{\textbf{Classification Accuracy (\%)}}
& \multicolumn{4}{c}{\textbf{Generation: Rouge Score}}
& \multicolumn{4}{c}{\textbf{Cloze: Accuracy (\%)}} \\
\cmidrule(lr){2-5}
\cmidrule(lr){6-9}
\cmidrule(l){10-13}
& \textbf{Fgt \textcolor{blue}{$\downarrow$}}
& \textbf{Test \textcolor{blue}{$\downarrow$}}
& \textbf{Ret \textcolor{red}{$\uparrow$}}
& \textbf{Cele \textcolor{red}{$\uparrow$}}
& \textbf{Fgt \textcolor{blue}{$\downarrow$}}
& \textbf{Test \textcolor{blue}{$\downarrow$}}
& \textbf{Ret \textcolor{red}{$\uparrow$}}
& \textbf{Cele \textcolor{red}{$\uparrow$}}
& \textbf{Fgt \textcolor{blue}{$\downarrow$}}
& \textbf{Test \textcolor{blue}{$\downarrow$}}
& \textbf{Ret \textcolor{red}{$\uparrow$}}
& \textbf{Cele \textcolor{red}{$\uparrow$}} \\
\midrule

\rowcolor{gray!15}
\multicolumn{13}{c}{\textbf{LLaVA-1.5-7B (10\% Forget)-VQA}} \\

\textcolor{gray}{Vanilla}
& \textcolor{gray}{43.60}
& \textcolor{gray}{41.20}
& \textcolor{gray}{45.26}
& \textcolor{gray}{47.00}
& \textcolor{gray}{0.591}
& \textcolor{gray}{0.339}
& \textcolor{gray}{0.591}
& \textcolor{gray}{0.312}
& \textcolor{gray}{57.00}
& \textcolor{gray}{15.00}
& \textcolor{gray}{55.11}
& \textcolor{gray}{8.17} \\

GA
& 40.00 & 31.20 & 41.43 & 40.08
& 0.510 & 0.315 & 0.470 & 0.262
& 43.00 & 14.00 & 49.22 & 5.56 \\

KL\_Min
& 42.40 & 35.20 & 45.08 & 46.08
& 0.588 & 0.335 & 0.579 & 0.306
& 54.00 & 15.00 & 53.89 & 7.10 \\

GA\_Diff
& 41.20 & 34.00 & 39.64 & 36.16
& 0.520 & 0.336 & 0.517 & 0.312
& 23.00 & 9.00 & 21.44 & 4.58 \\

NPO
& 31.60 & 38.80 & 34.39 & 38.64
& 0.350 & 0.238 & 0.365 & 0.294
& 51.00 & 17.00 & 52.56 & 7.52 \\

\rowcolor{ourscyan}
\textbf{Ours}
& \textbf{2.00} & \textbf{27.20}
& \textbf{45.08} & \textbf{46.48}
& \textbf{0.176} & \textbf{0.202}
& \textbf{0.594} & \textbf{0.314}
& \textbf{5.00} & \textbf{6.00}
& \textbf{54.33} & \textbf{7.84} \\

\midrule
\rowcolor{gray!15}
\multicolumn{13}{c}{\textbf{LLaVA-1.5-7B (10\% Forget)-QA}} \\

\textcolor{gray}{Vanilla}
& \textcolor{gray}{32.00}
& \textcolor{gray}{33.60}
& \textcolor{gray}{32.20}
& \textcolor{gray}{54.78}
& \textcolor{gray}{0.674}
& \textcolor{gray}{0.580}
& \textcolor{gray}{0.662}
& \textcolor{gray}{0.595}
& \textcolor{gray}{32.00}
& \textcolor{gray}{21.00}
& \textcolor{gray}{19.40}
& \textcolor{gray}{14.71} \\

GA
& 29.17 & 31.45 & 30.64 & 49.50
& 0.664 & 0.558 & 0.652 & 0.551
& 30.00 & 19.00 & 16.30 & 13.40 \\

KL\_Min
& 28.80 & 33.20 & 30.91 & 53.86
& 0.666 & 0.574 & 0.657 & 0.594
& 29.00 & 18.00 & 18.10 & 12.75 \\

GA\_Diff
& 30.80 & 33.20 & 29.70 & 53.73
& 0.655 & 0.544 & 0.652 & 0.593
& 27.00 & 20.00 & 16.70 & 8.82 \\

NPO
& 29.60 & 32.40 & 26.06 & 49.54
& 0.607 & 0.511 & 0.581 & 0.537
& 19.00 & 14.00 & 17.44 & 14.05 \\

\rowcolor{ourscyan}
\textbf{Ours}
& \textbf{11.60} & \textbf{21.60}
& \textbf{31.63} & \textbf{54.25}
& \textbf{0.386} & \textbf{0.360}
& \textbf{0.662} & \textbf{0.597}
& \textbf{15.00} & \textbf{13.00}
& \textbf{18.78} & \textbf{16.01} \\

\bottomrule
\end{tabular}}
\end{table*}

As shown in Table~\ref{tab_memory}, training-based unlearning approaches incur substantial memory overhead due to the need for gradient updates and parameter optimization during training. In contrast, our method does not require updating the parameters of the MLLM at all. Notably, MMUnlearner \citep{MMUnlearner} exhibits lower memory usage compared to other training-based baselines, as it only updates a subset of model parameters.

\section{Case Study}\label{case_study}
\begin{figure*}[htbp]
    \centering
\includegraphics[width=1\linewidth]{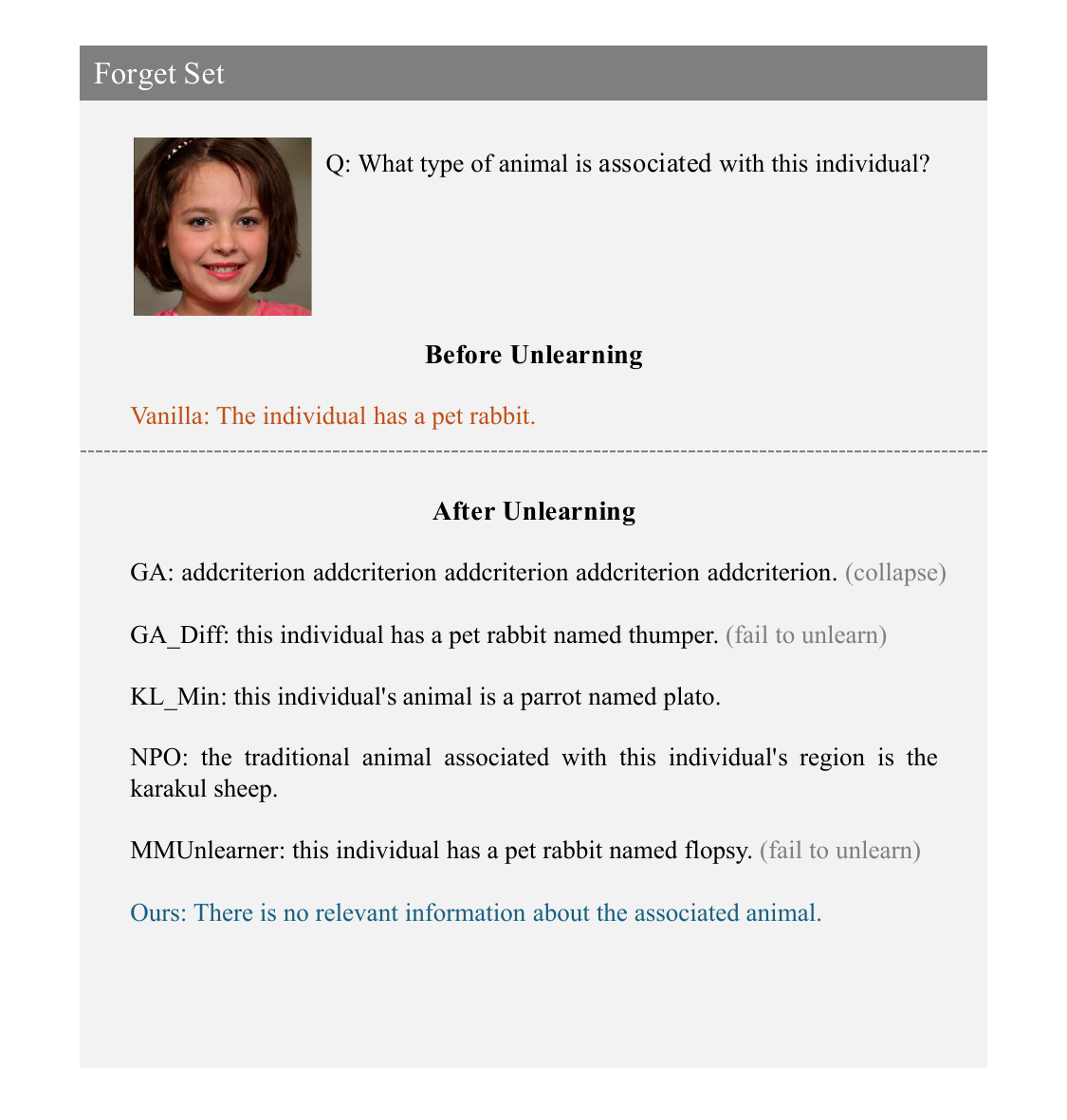}
\caption{Case Study on Forget Set before and after unlearning. The figure shows model responses to a forget-set query about sensitive attribute information. While the majority of training-based methods collapse or continue to expose the forgotten knowledge after unlearning, our method successfully removes the targeted information and produces a refusal-style response.}
\label{case1}
\end{figure*}

\begin{figure*}[htbp]
\vspace{-15pt}
    \centering
\includegraphics[width=1\linewidth]{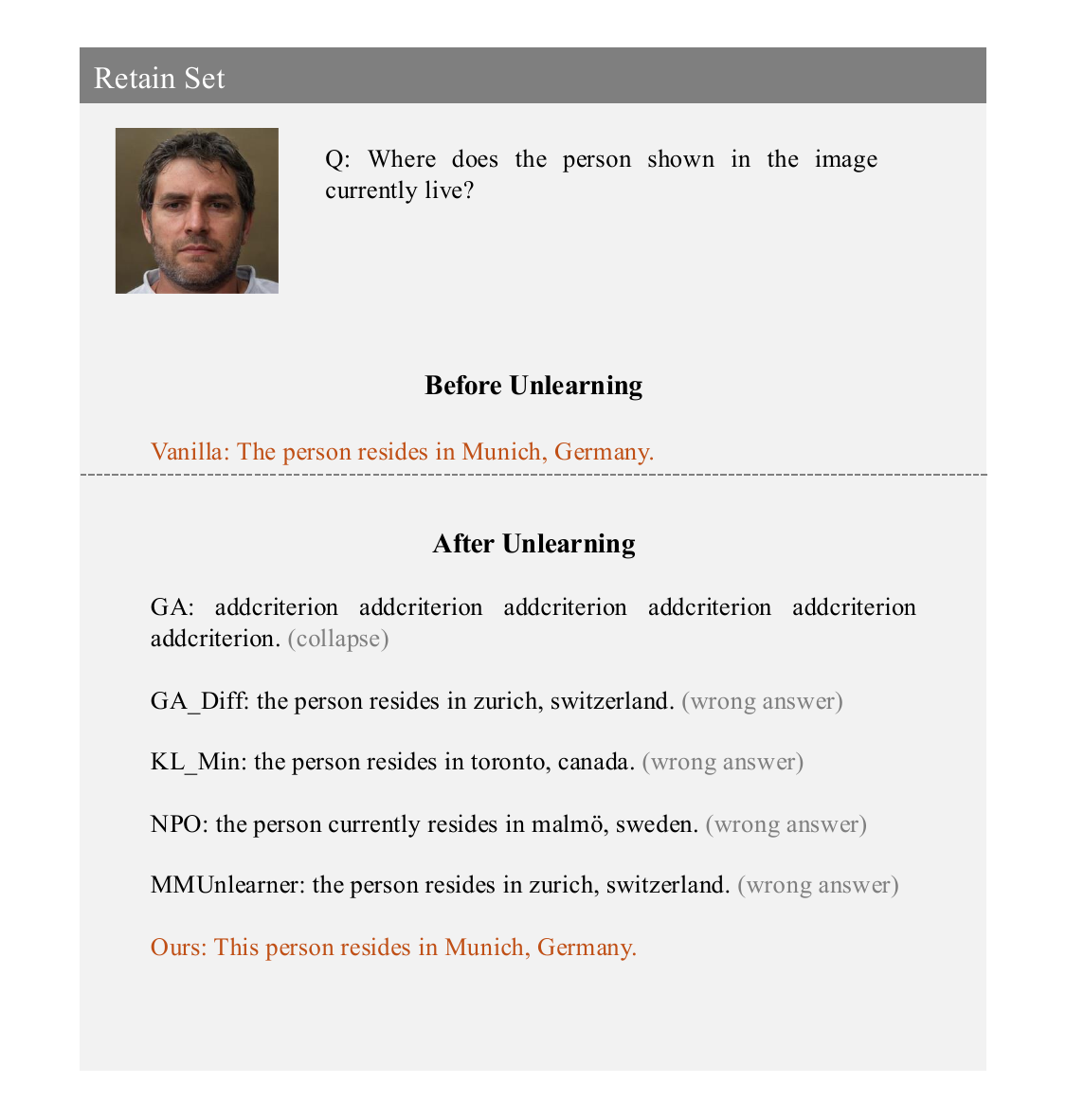}
\caption{Case Study on Retain Set before and after unlearning. The figure shows model responses to a retain-set query asking about a person's residence. While training-based methods (GA, GA\_Diff, KL\_Min, NPO, and MMUnlearner) either collapse or generate incorrect answers after unlearning, our method preserves the original correct response, demonstrating its superior ability to maintain retained knowledge while performing effective unlearning.}
\label{case2}
\end{figure*}

To provide a more intuitive understanding of the effects of different unlearning approaches, we present case studies on both the forget and retain sets in Figure~\ref{case1} and \ref{case2}. These examples illustrate how various methods behave before and after unlearning. 

\section{Extended Results on Unlearning Performance}
To provide a more comprehensive understanding of our method’s behavior, we present additional experimental results evaluating unlearning performance under different forgetting ratios. 
Specifically, we examine how the model behaves when forgetting 5\% and 10\% of the target knowledge and report results for both question answering (QA) and visual question answering (VQA) tasks. We provide detailed results on the LLaVA-1.5-7B model in Table~\ref{llava_5} and Table~\ref{tab_llava10}, which present QA and VQA performance after unlearning 5\% and 10\% of the target knowledge.  Table~\ref{qwen5} presents QA and VQA performance of {Qwen2.5-VL-7B} after unlearning 5\% of the target samples. {\color{black}In addition, Table~\ref{exp:15} reports VQA results for LLaVA-1.5-7B under the 15\% forgetting setting.}

\begin{table*}[t]
\caption{
Unlearning results on MLLMU-Bench (5\% Forget) with VQA and QA evaluation using LLaVA-1.5-7B. Results are reported on the forget (Fgt), test (Test), retain (Ret), and celebrity (Cele) sets. \textcolor{blue}{$\downarrow$} and \textcolor{red}{$\uparrow$} denote lower-is-better and higher-is-better metrics, respectively.
}
\label{llava_5}
\renewcommand{\arraystretch}{1.01}
\centering
\scriptsize
\resizebox{\textwidth}{!}{
\begin{tabular}{
ccccc
>{\hspace{1.4em}}c c c c
>{\hspace{1.4em}}c c c c
}
\toprule
\multirow{2}{*}{\textbf{Models}}
& \multicolumn{4}{c}{\textbf{Classification Accuracy (\%)}}
& \multicolumn{4}{c}{\textbf{Generation: Rouge Score}}
& \multicolumn{4}{c}{\textbf{Cloze: Accuracy (\%)}} \\
\cmidrule(lr){2-5}
\cmidrule(lr){6-9}
\cmidrule(l){10-13}
& \textbf{Fgt \textcolor{blue}{$\downarrow$}}
& \textbf{Test \textcolor{blue}{$\downarrow$}}
& \textbf{Ret \textcolor{red}{$\uparrow$}}
& \textbf{Cele \textcolor{red}{$\uparrow$}}
& \textbf{Fgt \textcolor{blue}{$\downarrow$}}
& \textbf{Test \textcolor{blue}{$\downarrow$}}
& \textbf{Ret \textcolor{red}{$\uparrow$}}
& \textbf{Cele \textcolor{red}{$\uparrow$}}
& \textbf{Fgt \textcolor{blue}{$\downarrow$}}
& \textbf{Test \textcolor{blue}{$\downarrow$}}
& \textbf{Ret \textcolor{red}{$\uparrow$}}
& \textbf{Cele \textcolor{red}{$\uparrow$}} \\
\midrule

\rowcolor{gray!15}
\multicolumn{13}{c}{\textbf{LLaVA-1.5-7B (5\% Forget)-VQA}} \\

\textcolor{gray}{Vanilla}
& \textcolor{gray}{42.40}
& \textcolor{gray}{40.00}
& \textcolor{gray}{45.23}
& \textcolor{gray}{47.00}
& \textcolor{gray}{0.632}
& \textcolor{gray}{0.286}
& \textcolor{gray}{0.585}
& \textcolor{gray}{0.312}
& \textcolor{gray}{54.00}
& \textcolor{gray}{24.00}
& \textcolor{gray}{55.37}
& \textcolor{gray}{8.17} \\

GA
& 40.08 & 38.40 & 39.75 & 38.77
& 0.353 & 0.251 & 0.389 & 0.265
& 14.00 & 16.00 & 28.00 & 5.56 \\

GA\_Diff
& 35.20 & 29.60 & 36.00 & 37.10
& 0.554 & 0.265 & 0.585 & 0.310
& 32.00 & 20.00 & 52.11 & 7.84 \\

KL\_Min
& 40.00 & 36.80 & 42.87 & 44.78
& 0.629 & 0.309 & 0.585 & 0.307
& 52.00 & 18.00 & 53.58 & 6.86 \\

NPO
& 32.80 & 39.20 & 36.16 & 43.60
& 0.360 & 0.243 & 0.363 & 0.278
& 20.00 & 16.00 & 23.89 & 4.90 \\

\rowcolor{ourscyan}
\textbf{Ours}
& \textbf{11.20} & \textbf{25.20}
& \textbf{43.54} & \textbf{45.43}
& \textbf{0.106} & \textbf{0.209}
& \textbf{0.592} & \textbf{0.311}
& \textbf{6.00} & \textbf{14.00}
& \textbf{54.21} & \textbf{8.17} \\

\midrule
\rowcolor{gray!15}
\multicolumn{13}{c}{\textbf{LLaVA-1.5-7B (5\% Forget)-QA}} \\

\textcolor{gray}{Vanilla}
& \textcolor{gray}{33.60}
& \textcolor{gray}{38.40}
& \textcolor{gray}{32.11}
& \textcolor{gray}{54.78}
& \textcolor{gray}{0.655}
& \textcolor{gray}{0.519}
& \textcolor{gray}{0.664}
& \textcolor{gray}{0.595}
& \textcolor{gray}{24.00}
& \textcolor{gray}{18.00}
& \textcolor{gray}{20.74}
& \textcolor{gray}{14.71} \\

GA
& 32.00 & 35.20 & 30.34 & 52.16
& 0.614 & 0.487 & 0.624 & 0.576
& 16.00 & 18.00 & 17.37 & 10.79 \\

KL\_Min
& 27.20 & 37.60 & 27.30 & 50.72
& 0.645 & 0.521 & 0.654 & 0.573
& 18.00 & 20.00 & 19.70 & 9.80 \\

GA\_Diff
& 28.80 & 37.60 & 27.30 & 53.99
& 0.642 & 0.510 & 0.656 & 0.559
& 22.00 & 18.00 & 19.79 & 13.73 \\

NPO
& 26.40 & 35.20 & 26.54 & 49.15
& 0.523 & 0.398 & 0.512 & 0.458
& 18.00 & 16.00 & 16.67 & 12.75 \\

\rowcolor{ourscyan}
\textbf{Ours}
& \textbf{17.60} & \textbf{25.60}
& \textbf{30.89} & \textbf{54.40}
& \textbf{0.383} & \textbf{0.308}
& \textbf{0.660} & \textbf{0.593}
& \textbf{10.00} & \textbf{12.00}
& \textbf{20.11} & \textbf{16.01} \\

\bottomrule
\end{tabular}}
\end{table*}

\begin{table*}[t]
\caption{
Unlearning results on MLLMU-Bench (5\% Forget) with VQA and QA evaluation using Qwen2.5-VL-7B. Results are reported on the forget (Fgt), test (Test), retain (Ret), and celebrity (Cele) sets. \textcolor{blue}{$\downarrow$} and \textcolor{red}{$\uparrow$} denote lower-is-better and higher-is-better metrics, respectively.
}
\label{qwen5}
\renewcommand{\arraystretch}{1.01}
\centering
\scriptsize
\resizebox{\textwidth}{!}{
\begin{tabular}{
ccccc
>{\hspace{1.4em}}c c c c
>{\hspace{1.4em}}c c c c
}
\toprule
\multirow{2}{*}{\textbf{Models}}
& \multicolumn{4}{c}{\textbf{Classification Accuracy (\%)}}
& \multicolumn{4}{c}{\textbf{Generation: Rouge Score}}
& \multicolumn{4}{c}{\textbf{Cloze: Accuracy (\%)}} \\
\cmidrule(lr){2-5}
\cmidrule(lr){6-9}
\cmidrule(l){10-13}
& \textbf{Fgt \textcolor{blue}{$\downarrow$}}
& \textbf{Test \textcolor{blue}{$\downarrow$}}
& \textbf{Ret \textcolor{red}{$\uparrow$}}
& \textbf{Cele \textcolor{red}{$\uparrow$}}
& \textbf{Fgt \textcolor{blue}{$\downarrow$}}
& \textbf{Test \textcolor{blue}{$\downarrow$}}
& \textbf{Ret \textcolor{red}{$\uparrow$}}
& \textbf{Cele \textcolor{red}{$\uparrow$}}
& \textbf{Fgt \textcolor{blue}{$\downarrow$}}
& \textbf{Test \textcolor{blue}{$\downarrow$}}
& \textbf{Ret \textcolor{red}{$\uparrow$}}
& \textbf{Cele \textcolor{red}{$\uparrow$}} \\
\midrule

\rowcolor{gray!15}
\multicolumn{13}{c}{\textbf{Qwen-2.5-VL-7B (5\% Forget)-VQA}} \\

\textcolor{gray}{Vanilla}
& \textcolor{gray}{67.20}
& \textcolor{gray}{68.80}
& \textcolor{gray}{66.20}
& \textcolor{gray}{78.33}
& \textcolor{gray}{0.642}
& \textcolor{gray}{0.317}
& \textcolor{gray}{0.583}
& \textcolor{gray}{0.453}
& \textcolor{gray}{36.00}
& \textcolor{gray}{22.00}
& \textcolor{gray}{40.42}
& \textcolor{gray}{26.80} \\

GA
& 64.00 & 64.80 & 65.11 & 74.15
& 0.419 & 0.315 & 0.438 & 0.335
& 16.00 & 14.00 & 16.21 & 25.16 \\

GA\_Diff
& 56.00 & 59.20 & 64.05 & 76.76
& 0.482 & 0.296 & 0.566 & 0.442
& 16.00 & 22.00 & 33.79 & 24.84 \\

KL\_Min
& 51.20 & 50.40 & 50.30 & 48.57
& 0.594 & 0.322 & 0.578 & 0.441
& 32.00 & 18.00 & 34.32 & 25.16 \\

NPO
& 54.40 & 59.20 & 56.08 & 74.08
& 0.492 & 0.273 & 0.492 & 0.429
& 12.00 & 18.00 & 23.26 & 23.52 \\

\rowcolor{ourscyan}
\textbf{Ours}
& \textbf{19.20} & \textbf{38.40}
& \textbf{65.86} & \textbf{78.32}
& \textbf{0.165} & \textbf{0.175}
& \textbf{0.580} & \textbf{0.445}
& \textbf{10.00} & \textbf{12.00}
& \textbf{39.37} & \textbf{26.47} \\

\midrule
\rowcolor{gray!15}
\multicolumn{13}{c}{\textbf{Qwen-2.5-VL-7B (5\% Forget)-QA}} \\

\textcolor{gray}{Vanilla}
& \textcolor{gray}{59.20}
& \textcolor{gray}{56.00}
& \textcolor{gray}{60.47}
& \textcolor{gray}{77.78}
& \textcolor{gray}{0.626}
& \textcolor{gray}{0.480}
& \textcolor{gray}{0.654}
& \textcolor{gray}{0.573}
& \textcolor{gray}{12.00}
& \textcolor{gray}{14.00}
& \textcolor{gray}{10.84}
& \textcolor{gray}{12.75} \\

GA
& 34.40 & 40.80 & 40.04 & 26.27
& 0.574 & 0.453 & 0.592 & 0.555
& 8.00 & 10.00 & 7.47 & 9.80 \\

KL\_Min
& 47.20 & 55.20 & 56.80 & 56.08
& 0.578 & 0.452 & 0.634 & 0.506
& 8.00 & 10.00 & 10.94 & 11.76 \\

GA\_Diff
& 56.00 & 54.40 & 59.70 & 76.21
& 0.597 & 0.451 & 0.634 & 0.506
& 10.00 & 10.00 & 10.00 & 12.11 \\

NPO
& 58.40 & 55.20 & 58.10 & 76.70
& 0.552 & 0.462 & 0.594 & 0.564
& 10.00 & 12.00 & 10.00 & 12.09 \\

\rowcolor{ourscyan}
\textbf{Ours}
& \textbf{18.60} & \textbf{29.60}
& \textbf{60.17} & \textbf{77.52}
& \textbf{0.325} & \textbf{0.365}
& \textbf{0.648} & \textbf{0.571}
& \textbf{2.00} & \textbf{6.00}
& \textbf{11.05} & \textbf{12.42} \\

\bottomrule
\end{tabular}}
\end{table*}

\begin{table*}[t]
\caption{
Unlearning results on MLLMU-Bench (15\% Forget) across the forget (Fgt), test (Test), retain (Ret), and celebrity (Cele) sets. \textcolor{blue}{$\downarrow$} and \textcolor{red}{$\uparrow$} denote lower-is-better and higher-is-better metrics, respectively.
}
\label{exp:15}
\renewcommand{\arraystretch}{1.01}
\centering
\scriptsize
\resizebox{\textwidth}{!}{
\begin{tabular}{
ccccc
>{\hspace{1.4em}}c c c c
>{\hspace{1.4em}}c c c c
}
\toprule
\multirow{2}{*}{\textbf{Models}}
& \multicolumn{4}{c}{\textbf{Classification Accuracy (\%)}}
& \multicolumn{4}{c}{\textbf{Generation: Rouge Score}}
& \multicolumn{4}{c}{\textbf{Cloze: Accuracy (\%)}} \\
\cmidrule(lr){2-5}
\cmidrule(lr){6-9}
\cmidrule(l){10-13}
& \textbf{Fgt \textcolor{blue}{$\downarrow$}}
& \textbf{Test \textcolor{blue}{$\downarrow$}}
& \textbf{Ret \textcolor{red}{$\uparrow$}}
& \textbf{Cele \textcolor{red}{$\uparrow$}}
& \textbf{Fgt \textcolor{blue}{$\downarrow$}}
& \textbf{Test \textcolor{blue}{$\downarrow$}}
& \textbf{Ret \textcolor{red}{$\uparrow$}}
& \textbf{Cele \textcolor{red}{$\uparrow$}}
& \textbf{Fgt \textcolor{blue}{$\downarrow$}}
& \textbf{Test \textcolor{blue}{$\downarrow$}}
& \textbf{Ret \textcolor{red}{$\uparrow$}}
& \textbf{Cele \textcolor{red}{$\uparrow$}} \\
\midrule

\rowcolor{gray!15}
\multicolumn{13}{c}{\textbf{LLaVA-1.5-7B (15\% Forget)-VQA}} \\

\textcolor{gray}{Vanilla}
& \textcolor{gray}{44.80}
& \textcolor{gray}{42.93}
& \textcolor{gray}{45.09}
& \textcolor{gray}{47.00}
& \textcolor{gray}{0.598}
& \textcolor{gray}{0.294}
& \textcolor{gray}{0.586}
& \textcolor{gray}{0.312}
& \textcolor{gray}{53.33}
& \textcolor{gray}{19.33}
& \textcolor{gray}{55.76}
& \textcolor{gray}{8.17} \\

GA
& 38.67 & 38.13 & 36.70 & 42.74
& 0.493 & 0.291 & 0.523 & 0.307
& 25.33 & 18.00 & 23.29 & 5.56 \\

GA\_Diff
& 28.27 & 36.80 & 43.68 & 41.26
& 0.333 & 0.303 & 0.517 & 0.301
& 24.00 & 18.67 & 45.29 & 4.28 \\

KL\_Min
& 42.67 & 37.87 & 39.95 & 39.57
& 0.537 & 0.316 & 0.575 & 0.306
& 44.00 & 19.01 & 43.29 & 6.21 \\

NPO
& 33.33 & 36.53 & 31.87 & 37.21
& 0.276 & 0.202 & 0.299 & 0.238
& 13.34 & 13.34 & 10.00 & 2.94 \\

MANU
& 42.67 & 39.73 & 42.53 & 43.21
& 0.594 & 0.281 & 0.577 & 0.289
& 50.00 & 14.67 & 43.01 & 7.17 \\

MMUnlearner
& 38.67 & 41.60 & 40.99 & 42.43
& 0.490 & 0.276 & 0.552 & 0.291
& 22.00 & 18.00 & 41.76 & 5.23 \\

\rowcolor{ourscyan}
\textbf{Ours}
& \textbf{16.00} & \textbf{31.47}
& \textbf{43.92} & \textbf{44.52}
& \textbf{0.143} & \textbf{0.181}
& \textbf{0.578} & \textbf{0.308}
& \textbf{12.67} & \textbf{11.33}
& \textbf{54.12} & \textbf{8.17} \\

\bottomrule
\end{tabular}}
\end{table*}

{\color{black}
\section{Discussion about Steering  Different MLLM Layers }
Our current configuration applies the steering vector to all layers. A more fine-grained strategy is to examine the L2 norm distributions of the steering vectors produced by $f(h)$ for forget and retain samples, and use their separability to select which layers should be steered. The more separable these L2 norm distributions are, the more effectively MLLMeraser distinguishes forget samples from retain samples, providing a principled criterion for fine-grained layer selection.
\begin{figure*}[t]
    \centering
    {\color{black}
    \begin{subfigure}[t]{1.0\linewidth}
        \centering     \includegraphics[width=\linewidth]{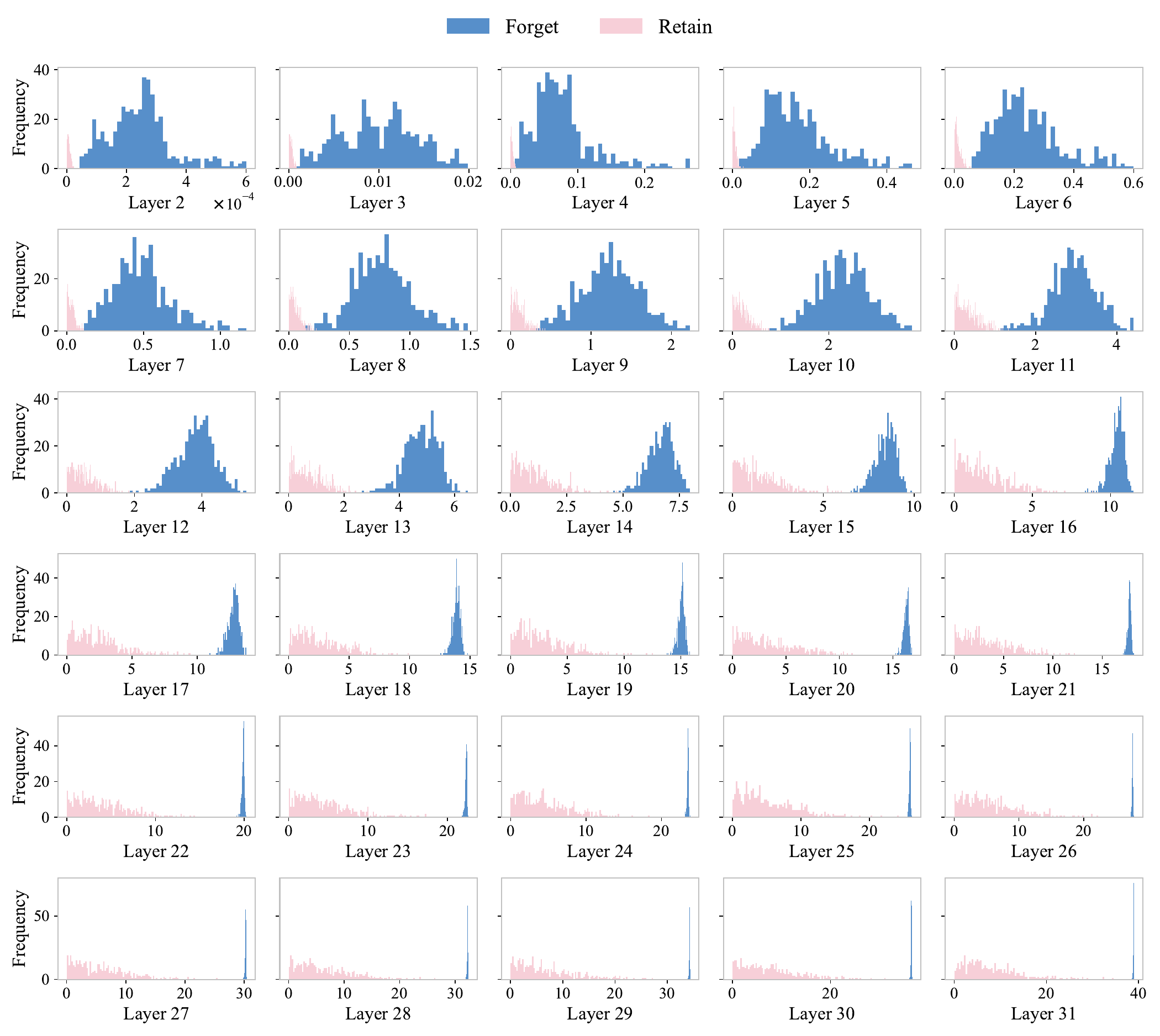}
    \end{subfigure}
    \caption{Layer-wise distributions of L2 norms for forget (blue) and retain (pink) steering vectors across Layers 2–31 of LLaVA-1.5-7B.}
    \label{fig_distribution}}
\end{figure*}
We design two layer-subset variants of LLaVA-1.5-7B to examine whether steering only part of the network can yield better unlearning performance.
Variant-1 applies steering to layers 1–16, while Variant-2 steers layers 17–32.
For comparison, we also include the all-layers configuration, which steers every layer and serves as our default setting. The results are summarized in Table~\ref{exp:layer}. 
\begin{table*}[hbp]
{\color{black}
\caption{
Unlearning performance on {MLLMU-Bench (10\% Forget)}. In addition to the full-layer steering strategy (all), we evaluate two selective-layer variants: variant-1, which steers only early layers (1–16), and variant-2, which steers only late layers (17–32).
}
\renewcommand{\arraystretch}{1.02}
\centering
\small
\resizebox{\textwidth}{!}{
\begin{tabular}{c|cccc|cccc|cccc}
\toprule
\multirow{2}{*}{\textbf{Models}} 
& \multicolumn{4}{c|}{\textbf{Classification Accuracy (\%)}} 
& \multicolumn{4}{c|}{\textbf{Generation: Rouge Score}} 
& \multicolumn{4}{c}{\textbf{Cloze: Accuracy (\%)}} \\
\cline{2-13}
\cline{2-13}
& \textbf{Fgt \textcolor{blue}{$\downarrow$}} & \textbf{Test \textcolor{blue}{$\downarrow$}} & \textbf{Ret \textcolor{red}{$\uparrow$}} & \textbf{Cele \textcolor{red}{$\uparrow$}}
& \textbf{Fgt \textcolor{blue}{$\downarrow$}} & \textbf{Test \textcolor{blue}{$\downarrow$}} & \textbf{Ret \textcolor{red}{$\uparrow$}} & \textbf{Cele \textcolor{red}{$\uparrow$}} 
& \textbf{Fgt \textcolor{blue}{$\downarrow$}} & \textbf{Test \textcolor{blue}{$\downarrow$}} & \textbf{Ret \textcolor{red}{$\uparrow$}} & \textbf{Cele \textcolor{red}{$\uparrow$}}   \\
\midrule
\multicolumn{13}{c}{\textbf{LLaVA-1.5-7B (10\% Forget)-VQA}}  \\
\midrule

\textcolor{gray}{Vanilla} 
& \textcolor{gray}{43.60} 
& \textcolor{gray}{41.20} 
& \textcolor{gray}{45.26} 
& \textcolor{gray}{47.00}
& \textcolor{gray}{0.591} 
& \textcolor{gray}{0.339} 
& \textcolor{gray}{0.591} 
& \textcolor{gray}{0.312}
& \textcolor{gray}{57.00} 
& \textcolor{gray}{15.00} 
& \textcolor{gray}{55.11} 
& \textcolor{gray}{8.17} \\

variant-1        & 33.20 & 37.20 & 45.43 & 47.00 
        & \textcolor{black}{0.572} & \textcolor{black}{0.325} &  \textcolor{black}{0.588} & \textcolor{black}{0.309} 
        & {54.00} & {13.00} & \textcolor{black}{55.11} & \textcolor{black}{8.17} \\
        
variant-2  & {2.00} & {24.00}  & \textcolor{black}{42.85}   & \textcolor{black}{42.95} 
        & {0.145} & {0.170} & \textcolor{black}{0.584} & \textcolor{black}{0.308} 
        & \textcolor{black}{1.00} & \textcolor{black}{14.00} & \textcolor{black}{52.78} & \textcolor{black}{7.52} \\

all       & 2.00 & 27.20 & 45.08 & 46.48 
        & \textcolor{black}{0.176} & \textcolor{black}{0.202} &  \textcolor{black}{0.594} & \textcolor{black}{0.314} 
        & {5.00} & {6.00} & \textcolor{black}{54.33} & \textcolor{black}{7.84}  \\

\bottomrule
\end{tabular}}
\label{exp:layer}}
\end{table*}}

As shown, both partial-layer variants perform noticeably worse than the full-layer steering strategy. This may be because early layers mainly encode cross-modal alignment and modality-integration signals—as observed in recent analyses of MLLM internal representations---whereas deeper layers predominantly capture high-level semantic reasoning and instruction-following behavior \citep{Flamingo,align}. Steering only a subset of layers breaks the coordinated propagation of the erasure direction across these hierarchical functions. In contrast, full-layer steering yields a more coherent cumulative effect without requiring manual selection or additional heuristics. Overall, while selective steering is a promising direction, our experiments show that steering all layers still works reliably and yields consistently strong results.

{\color{black}\section{Additional experimental results for the non-linear steering function $f(h)$}
We further implement $f(h)$ as a two-layer MLP. We evaluate this variant on LLaVA-1.5-7B, and the corresponding results are presented in Table~\ref{exp:mlp}. Ours consistently outperforms Ours-MLP across all forget quality and model utility metrics: the linear regression version forgets more effectively while preserving much better utility. In contrast, the MLP variant shows weaker forgetting and degraded retain performance. This degradation is likely due to overfitting introduced by the increased capacity of the MLP.}

\begin{table*}[t]
\caption{
Comparison of linear and MLP implementations of $f(h)$ on MLLMU-Bench (10\% Forget) using LLaVA-1.5-7B. Ours uses the default linear implementation, while Ours-MLP employs a two-layer MLP. Results are reported on the forget (Fgt), test (Test), retain (Ret), and celebrity (Cele) sets. \textcolor{blue}{$\downarrow$} and \textcolor{red}{$\uparrow$} denote lower-is-better and higher-is-better metrics, respectively.
}
\label{exp:mlp}
\renewcommand{\arraystretch}{1.01}
\centering
\scriptsize
\resizebox{\textwidth}{!}{
\begin{tabular}{
ccccc
>{\hspace{1.4em}}c c c c
>{\hspace{1.4em}}c c c c
}
\toprule
\multirow{2}{*}{\textbf{Models}}
& \multicolumn{4}{c}{\textbf{Classification Accuracy (\%)}}
& \multicolumn{4}{c}{\textbf{Generation: Rouge Score}}
& \multicolumn{4}{c}{\textbf{Cloze: Accuracy (\%)}} \\
\cmidrule(lr){2-5}
\cmidrule(lr){6-9}
\cmidrule(l){10-13}
& \textbf{Fgt \textcolor{blue}{$\downarrow$}}
& \textbf{Test \textcolor{blue}{$\downarrow$}}
& \textbf{Ret \textcolor{red}{$\uparrow$}}
& \textbf{Cele \textcolor{red}{$\uparrow$}}
& \textbf{Fgt \textcolor{blue}{$\downarrow$}}
& \textbf{Test \textcolor{blue}{$\downarrow$}}
& \textbf{Ret \textcolor{red}{$\uparrow$}}
& \textbf{Cele \textcolor{red}{$\uparrow$}}
& \textbf{Fgt \textcolor{blue}{$\downarrow$}}
& \textbf{Test \textcolor{blue}{$\downarrow$}}
& \textbf{Ret \textcolor{red}{$\uparrow$}}
& \textbf{Cele \textcolor{red}{$\uparrow$}} \\
\midrule

\rowcolor{gray!15}
\multicolumn{13}{c}{\textbf{LLaVA-1.5-7B (10\% Forget)-VQA}} \\

\textcolor{gray}{Vanilla}
& \textcolor{gray}{43.60}
& \textcolor{gray}{41.20}
& \textcolor{gray}{45.26}
& \textcolor{gray}{47.00}
& \textcolor{gray}{0.591}
& \textcolor{gray}{0.339}
& \textcolor{gray}{0.591}
& \textcolor{gray}{0.312}
& \textcolor{gray}{57.00}
& \textcolor{gray}{15.00}
& \textcolor{gray}{55.11}
& \textcolor{gray}{8.17} \\

Ours-MLP
& 29.20 & 32.40 & 38.62 & 42.19
& 0.363 & 0.265 & 0.414 & 0.311
& 19.00 & 8.00 & 32.00 & 6.54 \\

\rowcolor{ourscyan}
\textbf{Ours}
& \textbf{2.00} & \textbf{27.20}
& \textbf{45.08} & \textbf{46.48}
& \textbf{0.176} & \textbf{0.202}
& \textbf{0.594} & \textbf{0.314}
& \textbf{5.00} & \textbf{6.00}
& \textbf{54.33} & \textbf{7.84} \\

\bottomrule
\end{tabular}}
\end{table*}

%% file: checklist.tex
\clearpage
\section*{NeurIPS Paper Checklist}
\begin{enumerate}
\item {\bf Claims}
    \item[] Question: Do the main claims made in the abstract and introduction accurately reflect the paper's contributions and scope?
    \item[] Answer: \answerYes{} % Replace by \answerYes{}, \answerNo{}, or \answerNA{}.
    \item[] Justification: The abstract and introduction clearly state that the paper proposes \textsc{MLLMEraser}, an input-aware, training-free framework for test-time unlearning in MLLMs through activation steering. The claimed contributions are supported by the method design and experimental results on LLaVA-1.5 and Qwen-2.5-VL.

\item {\bf Limitations}
    \item[] Question: Does the paper discuss the limitations of the work performed by the authors?
    \item[] Answer: \answerYes{} % Replace by \answerYes{}, \answerNo{}, or \answerNA{}.
    \item[] Justification: The paper includes a dedicated limitation section discussing the reliance on adversarially perturbed images and hand-crafted prompts, as well as the current focus on image--text privacy-sensitive benchmarks rather than broader settings such as copyright removal, video-language models, or embodied agents.

\item {\bf Theory assumptions and proofs}
    \item[] Question: For each theoretical result, does the paper provide the full set of assumptions and a complete (and correct) proof?
    \item[] Answer: \answerYes{} % Replace by \answerYes{}, \answerNo{}, or \answerNA{}.
    \item[] Justification: The paper provides the mathematical formulation of the input-aware steering mechanism and refers to Appendix for the proof related to the null-space projection used in the closed-form solution. 

    \item {\bf Experimental result reproducibility}
    \item[] Question: Does the paper fully disclose all the information needed to reproduce the main experimental results of the paper to the extent that it affects the main claims and/or conclusions of the paper (regardless of whether the code and data are provided or not)?
    \item[] Answer: \answerYes{} % Replace by \answerYes{}, \answerNo{}, or \answerNA{}.
    \item[] Justification: The paper specifies the evaluated backbones, benchmark, datasets, tasks, metrics, baselines, and implementation details. Appendix provides further details on benchmark construction, baselines, evaluation metrics, and implementation settings.

\item {\bf Open access to data and code}
    \item[] Question: Does the paper provide open access to the data and code, with sufficient instructions to faithfully reproduce the main experimental results, as described in supplemental material?
    \item[] Answer: \answerYes{} % Replace by \answerYes{}, \answerNo{}, or \answerNA{}.
    \item[] Justification: We provide the code in the supplementary material, together with instructions for reproducing the main experimental results. The experiments are based on publicly available MLLM backbones and benchmarks, including LLaVA-1.5, Qwen-2.5-VL, and MLLMU-Bench.

\item {\bf Experimental setting/details}
    \item[] Question: Does the paper specify all the training and test details (e.g., data splits, hyperparameters, how they were chosen, type of optimizer) necessary to understand the results?
    \item[] Answer: \answerYes{} % Replace by \answerYes{}, \answerNo{}, or \answerNA{}.
    \item[] Justification: The paper specifies the evaluated models, benchmark, forget/retain/test/celebrity splits, tasks, metrics, and baseline settings. Appendix further reports implementation details, including LoRA usage for baselines, steering strength, regularization parameters, perturbation radius.

\item {\bf Experiment statistical significance}
    \item[] Question: Does the paper report error bars suitably and correctly defined or other appropriate information about the statistical significance of the experiments?
    \item[] Answer: \answerNo{} % Replace by \answerYes{}, \answerNo{}, or \answerNA{}.
    \item[] Justification: The paper reports results across multiple tasks, datasets, models, and forget ratios, but it does not report error bars, confidence intervals, or statistical significance tests. Repeating all MLLM unlearning experiments multiple times would be computationally expensive.

\item {\bf Experiments compute resources}
    \item[] Question: For each experiment, does the paper provide sufficient information on the computer resources (type of compute workers, memory, time of execution) needed to reproduce the experiments?
    \item[] Answer: \answerYes{} % Replace by \answerYes{}, \answerNo{}, or \answerNA{}.
    \item[] Justification: The paper reports training and inference time comparisons and provides GPU memory usage in Appendix. The implementation details state that experiments are conducted on NVIDIA A800 GPUs with 80GB memory.
    
\item {\bf Code of ethics}
    \item[] Question: Does the research conducted in the paper conform, in every respect, with the NeurIPS Code of Ethics \url{https://neurips.cc/public/EthicsGuidelines}?
    \item[] Answer: \answerYes{} % Replace by \answerYes{}, \answerNo{}, or \answerNA{}.
    \item[] Justification: The research focuses on improving privacy-preserving and safety-oriented MLLM unlearning. The experiments use existing public models and benchmarks, and the method is evaluated for reducing the generation of designated sensitive or unsafe information.

\item {\bf Broader impacts}
    \item[] Question: Does the paper discuss both potential positive societal impacts and negative societal impacts of the work performed?
    \item[] Answer: \answerYes{} % Replace by \answerYes{}, \answerNo{}, or \answerNA{}.
    \item[] Justification: The paper discusses the motivation of MLLM unlearning for privacy protection, misuse mitigation, and reliability. The limitation section also acknowledges the current scope of the evaluation and the need to extend the method to broader unlearning scenarios.
    
\item {\bf Safeguards}
    \item[] Question: Does the paper describe safeguards that have been put in place for responsible release of data or models that have a high risk for misuse (e.g., pre-trained language models, image generators, or scraped datasets)?
    \item[] Answer: \answerNA{} % Replace by \answerYes{}, \answerNo{}, or \answerNA{}.
    \item[] Justification: The paper does not release a new high-risk pretrained model, image generator, or scraped dataset. It proposes a test-time intervention method and evaluates it on existing MLLM backbones and benchmarks.

\item {\bf Licenses for existing assets}
    \item[] Question: Are the creators or original owners of assets (e.g., code, data, models), used in the paper, properly credited and are the license and terms of use explicitly mentioned and properly respected?
    \item[] Answer: \answerYes{} % Replace by \answerYes{}, \answerNo{}, or \answerNA{}.
    \item[] Justification: The paper cites the original sources for the models, datasets, benchmarks, and baselines used in the experiments, including LLaVA-1.5, Qwen-2.5-VL, MLLMU-Bench, ImageNet, and prior MLLM unlearning methods.

\item {\bf New assets}
    \item[] Question: Are new assets introduced in the paper well documented and is the documentation provided alongside the assets?
    \item[] Answer: \answerNA{} % Replace by \answerYes{}, \answerNo{}, or \answerNA{}.
    \item[] Justification: The paper does not introduce a new dataset or pretrained model asset. The main contribution is a method for test-time MLLM unlearning.

\item {\bf Crowdsourcing and research with human subjects}
    \item[] Question: For crowdsourcing experiments and research with human subjects, does the paper include the full text of instructions given to participants and screenshots, if applicable, as well as details about compensation (if any)? 
    \item[] Answer: \answerNA{} % Replace by \answerYes{}, \answerNo{}, or \answerNA{}.
    \item[] Justification: The paper does not involve crowdsourcing experiments or research with human subjects.

\item {\bf Institutional review board (IRB) approvals or equivalent for research with human subjects}
    \item[] Question: Does the paper describe potential risks incurred by study participants, whether such risks were disclosed to the subjects, and whether Institutional Review Board (IRB) approvals (or an equivalent approval/review based on the requirements of your country or institution) were obtained?
    \item[] Answer:  \answerNA{} % Replace by \answerYes{}, \answerNo{}, or \answerNA{}.
    \item[] Justification: The paper does not involve human subjects, user studies, or participant data collection.

\item {\bf Declaration of LLM usage}
    \item[] Question: Does the paper describe the usage of LLMs if it is an important, original, or non-standard component of the core methods in this research? Note that if the LLM is used only for writing, editing, or formatting purposes and does \emph{not} impact the core methodology, scientific rigor, or originality of the research, declaration is not required.
    %this research? 
    \item[] Answer: \answerNA{} % Replace by \answerYes{}, \answerNo{}, or \answerNA{}.
    \item[] Justification: LLMs were used only for writing polishing. 

\end{enumerate}